\documentclass[preprint]{imsart}

\RequirePackage[OT1]{fontenc}
\RequirePackage{amsthm,amsmath}
\RequirePackage[numbers]{natbib}
\RequirePackage[colorlinks,citecolor=blue,urlcolor=blue]{hyperref}
\usepackage{graphicx}
\usepackage[figuresright]{rotating}
\usepackage{amsmath}
\usepackage{times}
\usepackage{amsfonts,amsbsy, amssymb,float, mathtools,multirow, url}
\usepackage{bm}
\newcommand{\bs}[1] {\bm{#1}}
\usepackage{natbib}
\usepackage{comment}
\usepackage{enumitem}
\usepackage[ruled, lined, linesnumbered, commentsnumbered, longend]{algorithm2e}
\usepackage{xcolor}
\startlocaldefs
\def\bSig\mathbf{\Sigma}

\DeclareMathOperator*{\E}{\rm E}

\def\bs{\boldsymbol}

\numberwithin{equation}{section}
\theoremstyle{plain}
\newtheorem{theorem}{Theorem}[section]

\newtheorem{remark}{Remark}
\newtheorem{Assumption}{Assumption}
\endlocaldefs 
\begin{document}
\begin{frontmatter}

\title{Multiclass classification for multidimensional functional data through deep neural networks}
\runtitle{Multiclass classification for multidimensional functional data}
\begin{aug}
\author{\fnms{Shuoyang} \snm{Wang} \ead[label=e1]{shuoyang.wang@yale.edu}}
\and
\author{\fnms{Guanqun} \snm{Cao} \thanksref{t2}\ead[label=e2]{gzc0009@auburn.edu}}
\and
\author {for the Alzheimer's
Disease Neuroimaging Initiative}
\address{
Department of Biostatistics, Yale University, U.S.A. \\
\printead{e1}
}
\address{ 
Department of Mathematics and Statistics, Auburn University, U.S.A. \\
\printead{e2}}

\thankstext{t2}{Cao's  research was also partially supported by the Simons Foundation under Grant \#849413.}
\runauthor{Wang et al.}

\end{aug}

\begin{abstract}
     The intrinsically infinite-dimensional features of the functional observations over multidimensional domains  render  the standard classification methods effectively inapplicable. To address this problem, we introduce  a novel multiclass functional deep neural network (mfDNN) classifier as an innovative data mining and classification tool.  Specifically,  we   consider sparse deep neural network architecture with rectifier linear unit (ReLU) activation function and  minimize the cross-entropy loss in the multiclass classification setup. This neural network architecture allows us to employ modern computational tools in the implementation. The convergence rates of the misclassification risk functions are also derived for both fully observed and discretely observed multidimensional functional data. We demonstrate the performance of mfDNN on simulated data and several benchmark datasets from different application domains.
\end{abstract}


\end{frontmatter}

\section{Introduction}
 Functional data classification has wide applications in many areas such as machine learning and artificial intelligence~\cite{Song:etal:08,Leng:Muller:06,Rossi:05,Chamroukhi:12}.   While the majority of the  work on functional data classification focuses on one-dimensional functional data cases, for example, temperature data over a certain period \citep{Ramsay:Silverman:05}, and speech recognition data (log-periodogram data) \citep{Li:Ghosal:18,wang:etal:21a}, and there are few results obtained for multidimensional functional data classification, for example, 2D or 3D images classification.  To the best of our knowledge, there is only one recent work \cite{wang:etal:22} where binary classification for general non-Gaussian multidimensional functional data was considered.  However, their proposed classifier is designed for fully observed functional data and only able to conduct binary classification.     
There is also a lack of literature  working on  multiclass classification for functional data. \cite{Li:Ghosal:18} investigated a Bayesian approach to estimate parameters in multiclass functional models.   
However, their method is still only suitable for fully observed one-dimensional functional data classification.

The availability of massive databases resulted in the development of scalable  machine learning methods to process  these data. In this paper, we are interested in multiclass classification for multidimensional functional data, including but not limited to 2D and 3D imaging data classification, in the framework of DNN. The existing statistical and machine learning methods belong to the general framework of multivariate analysis. That is, data are treated as vectors of discrete samples and permutation of components will not affect the analysis results, hence the ordering of the pixels or voxels is irrelevant in these analyses and the imaging data were treated as high-dimensional data. In recent years, many sparse  discriminant analysis methods have been proposed for i.i.d.
high-dimensional data classification and variable selection. Most of these proposals focus
on binary classification and they are not directly applicable to multiclass classification problems. A popular multiclass sparse discriminant analysis proposal  is the $\ell_1$ penalized Fisher’s
discriminant \cite{Witten:Tibshirani:11}. However,  \cite{Witten:Tibshirani:11} does not have theoretical justifications. It is generally unknown how close their estimated discriminant
directions are to the true directions, and whether the final classifier will work similarly as the Bayes
rule.  More recently, \cite{Mai:19} proposed a   multiclass sparse discriminant analysis method that estimates all discriminant directions
simultaneously with theoretical justification for high-dimensional data setting.   We  compare our method with both \cite{Witten:Tibshirani:11} and  \cite{Mai:19} carefully on   numerical performance  in Sections  \ref{SEC:simulation} and \ref{SEC:realdata}. 

An efficient way  to handle multi-dimensional functional data, for example, imaging data, is to incorporate the information that is inherent in order and smoothness of processes over pixels or voxels. 
 In this paper, we propose a novel approach, called as multi-class functional deep neural network (mfDNN, multiclass-functional+DNN), for multidimensional  functional data multiclass classification.  We  extract the functional principal components scores (FPCs) of the data functions, and then train a DNN based classifier on these FPCs as well as their corresponding class memberships. Given these FPCs as inputs and  their corresponding 
 class labels as outputs, we consider sparse deep ReLU network architecture and minimize the
cross-entropy (CE) loss in the multiclass classification setup. CE loss is a standard tool in machine learning classification,
but the technique has not been utilized to its fullest potential in the context of
functional data classification,  especially regarding the study of theoretical properties of DNN classifiers.
A recent work \cite{Bos:etal:21},  has an example of multiclassification by minimizing CE loss in sparse DNN for cross-sectional data setting.  While the
powerful capabilities of DNN and CE loss have been clearly demonstrated for i.i.d.  data, how to best adapt these models to functional data remains under-explored.  We apply the same loss function in \cite{Bos:etal:21} but develop it
in the context of multidimensional functional multiclassification. Furthermore, 
 convergence rates and   the conditional
class probabilities are developed under  both fully observed and discretely observed multi-dimensional functional data setting, respectively.

The rest of this article is organized as follows. In Section \ref{SEC:method}, we introduce the   $K$-class functional data classification models and propose the multiclass functional deep neural network classifiers.  
In Section \ref{SEC:theory}, we establish
theoretical properties of  mfDNN under suitable technical assumptions. Section \ref{SEC:Examples} provides two progressive examples to demonstrate the validity of these technical assumptions. In Section \ref{SEC:simulation}, performances of 
 mfDNN and its competitors are demonstrated through simulation studies. The corresponding R programs for implementing our method are provided on
GitHub. 
In Section \ref{SEC:realdata}, we apply  mfDNN to the zip code dataset and Alzeheimer’s Disease data. Section \ref{SEC:Summary}
summarizes the conclusions. Technical proofs are provided in Appendix. 


\section{Methodology} \label{SEC:method}
\subsection{$K$-class functional data classification models}
Suppose we observe $n$ i.i.d. functional training samples $\{(X_i (s), \bm Y_i): 1\le i\le n, s\in\left[ 0, 1\right]^d\}$,
which are independent of $X(s)$ to be classified, and class label $\bm Y_i = \left(Y_{i1}, \ldots, Y_{iK} \right)^\intercal$, such that $Y_{ik}=1$ if the sample is in the $k$-th group, $k=1,2,\ldots,K$, and $0$ otherwise.
Throughout the entire paper, we consider the number of class $K$ is a universal constant not depending on sample size $n$. In multiclass classification with $K \geq2$ classes,  we are interested in grouping a newly observed
continuous curves $X(s)$  to one of the $K$ classes given the training data.

 If $Y_{ik}=1$, $X_i(s)$ presumably has some unknown mean function $\mathbb{E}X_i(s)=\mu_k(s)$  and unknown covariance function $\Omega_k(s,s')=\mathbb{E}\left[\left(X_i(s)-\mu_k(s)\right)\left(X_i(s')-\mu_k(s')\right)\right]$, for $s,s'\in\left[ 0, 1\right]^d$ and $1\leq k\leq K$. 
Suppose the covariance function $\Omega_k(\cdot,\cdot)$ satisfies the Karhunen–Lo\`{e}ve decomposition:
\begin{equation}\label{eqn:Omegak}
\Omega_k(s,s')=\sum_{j=1}^\infty\lambda_{kj}\psi_{kj}(s)\psi_{kj}(s'), s,s'\in\left[ 0, 1\right]^d,
\end{equation}
where $\psi_{kj}$, $j\ge1$ is an orthonormal basis of $L^2(\left[ 0, 1\right]^d)$ with respect to the usual $L^2$ inner product,
and $\lambda_{k1}\ge \lambda_{k2}\ge \cdots>0$ are nonincreasing positive eigenvalues.  
For any $X(s)\in L^2(\left[ 0,1\right]^d)$ under the $k$-th class, write $X(s)=\sum_{j=1}^\infty\xi_j\psi_{kj}(s)$, where
$\xi_j$'s are  pairwise uncorrelated random coefficients. By projecting the function space to vector space, we define the conditional probabilities 
$$\pi_{k}(\bm x) = {\rm P}(\bm Y = \bm e_k|\bm\xi=\bm x), ~~  k\in \{1,\ldots, K\},$$
where  $\bm e_k = \left(0,\ldots,0, {1}, 0,\ldots, 0\right)^\intercal$ is a $K$-dimensional standard basis vector denoting
the label of the k-th class is observed and $\bm{\xi} =(\xi_1 ,\xi_2 ,\ldots)^\intercal$ is an infinite dimension random vector. Notably, $\sum_{k=1}^{K}\pi_{k}(\bm x) = 1$ for any   $\bm x \in \mathbb{R}^{\infty}$.

\subsection{Multiclass functional deep neural network classifier}\label{sec:classifier}
For $k\in \{1,\ldots, K \}$, define sample covariance function
\[
\widehat{\Omega}_k(s,s')=\frac{1}{n_k}\sum_{i\in I_k}(X_i(s)-\bar{X}_k(s))(X_i(s')-\bar{X}_k(s')),\,\,\,\,
s,s'\in\left[ 0, 1\right]^d,
\]
where $I_k$ is the collection of $i$ such that $Y_{ik}=1$,
$n_k=|I_k|$ is the sample size, and $\bar{X}_k(s)= n_k^{-1}\sum_{i\in I_k}X_i(s)$
is the sample mean function of class $k$.
The Karhunen–Lo\'{e}ve decomposition for $\widehat{\Omega}_k$ is
\[
\widehat{\Omega}_k(s,s')=\sum_{j=1}^\infty\widehat{\lambda}_{kj}\widehat{\psi}_{kj}(s)\widehat{\psi}_{kj}(s'), s,s'\in\left[ 0, 1\right]^d.
\]
The sample data function $X_i(s)$, $i=1,\ldots,n$, under $Y_{ik}=1$, can be represented as
$
X_i(s)=\sum_{j=1}^\infty\widehat{\xi}_{ij}\widehat{\psi}_{kj}(s)$.
Intuitively, $\widehat{\bm\xi}^{(i)}:=(\widehat{\xi}_{i1},\widehat{\xi}_{i2},\ldots)$ is an estimator of $\bm{\xi}^{(i)}:=(\xi_{i1},\xi_{i2},\ldots)$, in which
$\xi_{ij}$'s are unobservable random coefficients of $X_i$ with respect to the population basis $\psi_{kj}$. Hence, it is natural to design classifiers based on $\widehat{\bm\xi}^{(i)}$'s. 

Let $\widehat{\bm\xi}^{(i)}_{J}=(\widehat{\xi}_{i1},\ldots,\widehat{\xi}_{iJ})^\intercal$ be the $J$-dimensional truncation of $\widehat{\bm\xi}^{(i)}$ for $i=1,\ldots,n$.
Denote $h_k(\cdot)$ as the conditional probability densities of $ \bm\xi$ given $\bm Y=\bm e_k$, such that $\pi_{k}(\bm x) = \frac{h_{k}{\rm P}(\bm Y = \bm e_{k})}{\sum_{\ell=1}^K h_{\ell}{\rm P}(\bm Y = \bm e_{\ell})}$.
When $X_i$'s are some random processes with complex structures, such as non-Gaussian processes,
one major challenge is the underlying complicated form of $\{h_k\}_{k=1}^K$ so that estimation of $\{\pi_k\}_{k=1}^K$ is typically difficult.
In this section, inspired by the rich approximation power of DNN, we propose a new classifier so-called mfDNN,  which can still approach Bayes classifiers closely even when $h_k$'s are non-Gaussian and complicated.

Let $\sigma (x)=(x)_+$ be the ReLU activation function, and $K$-dimensional softmax activation function $\bm\sigma^\ast (\bm y) = \left(\frac{\exp{(y_1)}}{\sum_{k=1}^K\exp{(y_k)}},  \ldots, \frac{\exp{(y_K)}}{\sum_{k=1}^K\exp{(y_k)}}\right)$ for $\bm y=(y_1, \ldots, y_K)\in \mathbb{R}^K$.
For any real vectors $\bm{V}=(v_1,\ldots,v_w)^\intercal$ and $\bm{z}=(z_1,\ldots,z_w)^\intercal$, define the shift activation function $\sigma_{\bm{V}}(\bm{z})=(\sigma(z_1-v_1),\ldots,\sigma(z_w-v_w))^\intercal$.  
For $L\ge1$, $\bm{p}=(p_1,\ldots,p_L)\in\mathbb{N}^L$, 
let $\mathcal{F}(L,J,\bm{p})$ denote the class of fully connected feedforward DNN with $J$ inputs, $L$ hidden layers and, for $l=1,\ldots,L$, $p_l$ nodes on the $l$-th hidden layer.
Equivalently, any $f\in\mathcal{F}(L,J,\bm{p})$ has an expression
 \begin{equation} \label{EQ:f}
f(\bm x) = \bm\sigma^\ast\mathbf{W}_L\sigma_{\bm{V}_L} \mathbf{W}_{L-1}\sigma_{\bm{V}_{L-1}}\ldots \mathbf{W}_1\sigma_{\bm{V}_1} \mathbf{W}_0\bm x, \,\,\,\, {\bm x}\in\mathbb{R}^J,
\end{equation}
where $\mathbf{W}_l\in\mathbb{R}^{p_{l+1}\times p_{l}}$, for $l=0,\ldots,L$, are weight matrices, $\bm{V}_l\in\mathbb{R}^{p_l}$, for $l=1,\ldots,L$,
are shift vectors. Here, we adopt the convention that $p_0=J$ and $p_{L+1}=1$. 

Due to the large capacity of the fully connected DNN class and to avoid overparameterization, we
 consider the following sparse DNN class:
\begin{eqnarray}\label{EQ:class}
&& \mathcal{F}(L, J, \bm{p},   s)
 =  \\
 &&\left\{ f\in \mathcal{F}(L, J, \bm{p}) :  \sum_{l=0}^L \| \mathbf{W}_l\|_0+ \| \mathbf{V}_l\|_0 \leq s, \max_{0\le l\le L}\| \mathbf{W}_l\|_{\infty}\vee \|\mathbf{V}_l\|_{\infty}\le 1\right\}, \nonumber
\end{eqnarray}
where $ \| \cdot\|_{\infty}$ denotes the maximum-entry norm of a matrix/vector, which is controlled by $1$, $\| \cdot\|_0$ denotes the number of non-zero elements of a matrix/vector, and $s$ controls the total number of active neurons.



Given the training data
$(\bm\xi^{(1)}_J, \bm Y_1),\ldots,(\bm\xi^{(n)}_J, \bm Y_n)$, let
\begin{equation}\label{hatf:FDNN}
\widehat{\bm f}_{\phi}(\cdot) = \left(\widehat{f}_{\phi}^{(1)}, \ldots, \widehat{f}_{\phi}^{(K)} \right)^\intercal = \arg\min_{\bm f\in \mathcal{F}(L, J, \bm{p}, s)} -\frac{1}{n}\sum_{i=1}^{n }\phi(\bm Y_i, \bm f(\widehat{\bm\xi}^{(i)}_{J})),
\end{equation}
where $\phi(\bm x_1, \bm x_2) = \bm x_1 ^\intercal \log\left( \bm x_2\right)$ denotes the CE loss function.

We then propose the following mfDNN classifier:
for $X\in L^2(\left[ 0, 1\right]^d)$,
\begin{equation}\label{eq:classifier}
    {\widehat{G}^{mfDNN}(X)}= k, \textit{~ if }\widehat{f}_{\phi}^{(k)}(\bm\xi_J)\geq \widehat{f}_{\phi}^{(k')}(\bm\xi_J), \textit{~ for all ~} 1\leq k'\neq k\leq K.
\end{equation}

The detailed implementation for the proposed mfDNN classifier is explained in Algorithm \ref{alg:mfdnn}. The tuning parameters include $J,L,\bm p$, and $s$, which theoretically contribute to the performance of classification. We discuss the related theories in Section \ref{SEC:theory}. In practice, we suggest the following data-splitting approach to select $(J, L, \bm{p}, s)$. Note that the sparsity is not directly applicable by assigning the number of active neurons. Instead, we use dropout technique to achieve sparse network. 

\begin{algorithm}
\caption{Training algorithm for mfDNN}\label{alg:mfdnn}

\KwIn{Collection of samples $\left\{ \{X_i(s_j)\}_{j=1}^m, \bm Y_i\right\}_{i=1}^n$, training index set $\mathcal{I}_1$, testing index set $\mathcal{I}_2$, $|\mathcal{I}_1|= \lfloor0.7n\rfloor$, $|\mathcal{I}_2|= \lfloor0.3n\rfloor$, candidates for number of projection scores $\left\{ J_0,\ldots,J_{N_1}\right\}$, 
candidates for depth $\left\{ L_0,\ldots,L_{N_2}\right\}$,
candidates for width $\left\{ p_0,\ldots,p_{N_3}\right\}$,
candidates for dropout rates $\left\{ s_0,\ldots,s_{N_4}\right\}$, number of epochs, batch size, learning rate} 
\KwOut{$\left\{ {\widehat{G}^{mfDNN}(X_i)}\right\}_{i=1}^n$} 
Extract $\widehat{\bm\xi}^{(i)}_{J}$ from $\left\{ X_i(s_j)\right\}_{i,j=1}^{n,m}$ by integration, and $J=\max\{J_0,\ldots, J_{N_1} \}$\\
For all candidates,\linebreak
(i) Train $\widehat{\bm f}_{J_{\ell_1},L_{\ell_2},p_{\ell_3},s_{\ell_4}}$ by Equation (\ref{hatf:FDNN}) with $\left( \{X_i(s_j)\}_{j=1}^m, \bm Y_i\right)_{i\in \mathcal{I}_1}$.\linebreak
(ii) Compute $\textrm{err}(\ell_1,\ell_2,\ell_3,\ell_4) :=\frac{1}{|\mathcal{I}_2|}\sum_{i\in\mathcal{I}_2}I(\widehat{\bm f}_{J_{\ell_1},L_{\ell_2},p_{\ell_3},s_{\ell_4}}\neq \bm Y_i)$ \\
Obtain $\left( J_{\ell_1^\ast},L_{\ell_2^\ast},p_{\ell_3^\ast},s_{\ell_4^\ast}\right) = \arg\min_{\ell_1,\ell_2,\ell_3,\ell_4}\textrm{err}(\ell_1,\ell_2,\ell_3,\ell_4)$\\
Train $\widehat{\bm f}_{ J_{\ell_1^\ast},L_{\ell_2^\ast},p_{\ell_3^\ast},s_{\ell_4^\ast}}$ by Equation (\ref{hatf:FDNN}) with $\left( \{X_i(s_j)\}_{j=1}^m, \bm Y_i\right)_{i=1}^n$\\
\end{algorithm}

\section{Theoretical properties} \label{SEC:theory}

\subsection{Functional data space}

\subsubsection{Assumptions for functional multi-classification setup}
To formulate our classification task, we first introduce two mild assumptions for $\pi_{k}(\cdot)$. Without further notice,  $c$, $C$, $C_1$, $C_2$, \ldots, represent some positive constants and can vary from line to line.
\begin{Assumption} (Boundary condition)\label{A1} 
There exist an absolute constant $C>0$ and some positive vector $\bm\alpha = (\alpha_1, \ldots, \alpha_K)^\intercal $ such that 
\begin{equation}
    {\rm P}\left(\pi_{k}(\bm\xi) \le x\right)\le Cx^{\alpha_k}, \;\;\;\;\;\;\forall x\in \left(0, 1\right], \forall k\in \{1,\ldots, K\}.
\end{equation}
\end{Assumption}
Assumption \ref{A1} describes the behaviour of $\pi_{k}$ around $0$. Trivially, when $\alpha_k=0$ for $k=1,\ldots,K$, the condition holds universally only if $C\geq 1$.  Similar conditions can be found for multi-class classification for multivariate data in \cite{Bos:etal:21} and binary classification in \cite{Mammen:etal:99} and \cite{Tsybakov:04}.

\begin{Assumption} (Uniform lower bound)\label{A3}
There exists a relatively small $\epsilon\in \left( 0,1\right)$, such that
$${\rm P}\left( \pi_{k}(\bm\xi)>\epsilon\right)=1$$
for all $k=1,\ldots, K$.
\end{Assumption}
Assumption \ref{A3} provides a uniform lower bound of $\pi_{k}$, indicating that $\pi_{k}$ is bounded away from $\epsilon$ in probability.

\begin{remark}
Assumptions \ref{A1} and \ref{A3}  both depict the characteristic of $\pi_{k}$ around zero in different aspects. Specifically, Assumption \ref{A1} controls the decay rate of the probability measure of $\left\{\pi_{k}(\bm\xi): 0\leq \pi_{k}(\bm\xi)\leq x \right\}$. Assumption \ref{A3} truncates the probability measure below $\epsilon$ to be zero. It can be seen that Assumptions \ref{A1} and \ref{A3} are closely related in certain circumstances. For example, given some $\epsilon>0$, Assumption \ref{A3} implies Assumption \ref{A1} for arbitrary $\bm\alpha$ and $C=\max_k \epsilon^{-\alpha_k}$, and Assumption \ref{A1} implies Assumption \ref{A3} when $C=0$ in a trivial case.  However, they are not equivalent in most scenarios when $C$,$\{\bm\alpha_k\}_{k=1}^K$, and $\epsilon$ are all universally positive constants.


\end{remark}

Essentially, functional data have intrinsically infinite dimension. Owing to this unique phenomenon, we introduce the following finite approximation condition for $\bm \pi$ . 
For any positive integer $J$, define $\bm\xi^{(J)}=(\xi_1,  \ldots, \xi_J)^\intercal$ and $\bm \pi_J(\bm x_J) = \left(\pi_1^{(J)}(\bm x_J), \ldots, \pi^{(J)}_K(\bm x_J)\right)^\intercal$, where $\pi^{(J)}_k(\bm x_J) = {\rm P}(\bm Y_k = \bm e_k|\bm\xi_J=\bm x_J)$, $k=1,\ldots, K$.

\begin{Assumption}(Approximation error of $\pi^{(J)}_k$)\label{A2}
There exist a constant $J_0\ge1$ and decreasing functions $\zeta(\cdot):\left[1, \infty\right)\to\mathbb{R}_+$ and $\Gamma(\cdot):\left[0, \infty\right)\to\mathbb{R}_+$,
with $\sup_{J\ge1}J^\varrho \zeta(J)<\infty$ for some $\varrho >0$ and $\int_0^\infty\Gamma(x)dx<\infty$, such that for any $J\ge J_0$, $k=1\ldots, K$ and $x>0$,
\begin{equation}
    {\rm P}\left(|\pi_{k}(\bm\xi)-\pi^{(J)}_k(\bm\xi_J) | \geq x\right) \leq \zeta(J)\Gamma(x).
\end{equation}
\end{Assumption}
Assumption \ref{A2} provides a uniform upper bound on the probability of $\pi_{k}$ differing from $\pi^{(J)}_k$ by at least $x$, which approaches zero if either $J$ or $x$ tends to infinity, implying that when dimension is reduced, there exists some large enough $J$, such that $\pi^{(J)}_k$ is an accurate approximation of $\pi_{k}$. All three assumptions can be verified in concrete examples included in Section \ref{SEC:Examples}.

\subsubsection{Conditional probability $\pi_k$ with complex structure}


In the following, we impose composition structures on $\pi_{k}(\cdot)$.
For $t\ge1$, a measurable subset $D\subset \mathbb{R}^t$, and constants $\beta,R>0$,
define 
\begin{eqnarray*}
 \mathcal{C}^\beta(D, R)&= &\left\{f:D\mapsto\mathbb{R}\big|\sum_{\bm{\alpha}:|\bm{\alpha}|<\beta} \| \partial^{\bm{\alpha}}f\|_{\infty}f\right.\\
&&\left. + \sum_{\bm{\alpha}:|\bm{\alpha}|=\lfloor{\beta}\rfloor}\sup_{\bm{z}, \bm{z}' \in D, \bm{z} \neq \bm{z}' } \frac{|\partial^{\bm{\alpha}}f(\bm{z}) - \partial^{\mathbf{\alpha}}f(\bm{z}') |}{\|\bm{z} - \bm{z}'\|_{\infty}^{\beta - \lfloor\beta\rfloor}} \leq R\right\},
\end{eqnarray*}
where $\partial^{\bm{\alpha}}$ = $\partial^{\alpha_1}\ldots\partial^{\alpha_t}$ denotes the
partial differential operator with multi-index $\bm{\alpha}$ = $(\alpha_1, \ldots, \alpha_t) \in \mathbb{N}^t$, $|\bm{\alpha}|=\alpha_1+\cdots+\alpha_t$.
Equivalently, $\mathcal{C}^\beta(D, C)$ is the ball of $\beta$-H\"{o}lder smooth functions 
on $D$ with radius $R$.
A function $f: \mathbb{R}^t\to\mathbb{R}$ is said to be locally 
$\beta$-H\"{o}lder smooth if 
for any $a,b\in\mathbb{R}$, there exists a constant $C$ (possibly depending on $a,b$) such that $f\in\mathcal{C}^\beta([a,b]^t,C)$.

For $q\ge0, J\ge 1$, let $d_0=J$ and $d_{q+1}=1$. For $\bm{d}=(d_1, \ldots, d_q)\in\mathbb{N}_+^{q}$,  $\bm{t}= (t_0, \ldots, t_q)\in\mathbb{N}_+^{q+1}$ with $t_u\le d_u$ for $u=0,\ldots, q$, 
$\bm{\beta} := (\beta_0, \ldots, \beta_q)\in\mathbb{R}_+^{q+1}$,
let $\mathcal{G}(q, J, \bm d, \bm t, \bm\beta)$ be the class of functions $g$ 
satisfying a modular expression
\begin{equation}\label{modular:expression}
g(\bm{z})=g_q \circ \cdots \circ g_0(\bm{z})\in \left(0, 1\right),\,\,\forall \bm{z}\in\mathbb{R}^{d_0},
\end{equation}
where $g_u=(g_{u1},\ldots,g_{ud_{u+1}}): \mathbb{R}^{d_u}\mapsto\mathbb{R}^{d_{u+1}}$
and $g_{uv}: \mathbb{R}^{t_u}\mapsto\mathbb{R}$ are locally $\beta_u$-H\"{o}lder smooth.
The $d_u$ arguments of $g_u$ are locally connected
in the sense that each component $g_{uv}$ only relies on $t_u (\le d_u)$ arguments. 
Similar structures have been considered by \cite{Schmidt:19, Bauer:Kohler:19,Liu:etal:2021,  wang2021stat,Bos:etal:21,Kim:NN:2021, hu2020arxiv}
in multivariate regression or classification to overcome high-dimensionality. Generalized additive model \citep{gam1990} and tensor product space ANOVA model \citep{lin2000aos}
are special cases; see \cite{Liu:etal:2021}.

 We define the class of $\bm h =\{h_1,\ldots,h_K\}$ as
\begin{align*}
\mathcal{H}\equiv &\mathcal{H}\left(K, \{q^{(k)}\}_{k=1}^{K}, \{\bm{d}^{(k)}\}_{k=1}^{K}, \{\bm{t}^{(k)}\}_{k=1}^{K}, \{\bm{\beta}^{(k)}\}_{k=1}^{K}, \{ \alpha_k\}_{k=1}^{K}, \{ \pi_k(\cdot)\}_{k=1}^{K}, \right.\\
&\left.\zeta(\cdot),\Gamma(\cdot), C, \rho, \epsilon\right),    
\end{align*}
such that for any $J\geq 1$, $\pi^{(J)}_k\in\mathcal{G}\left(q^{(k)}, J, \bm{d}^{(k)}, \bm{t}^{(k)}, \bm{\beta}^{(k)}\right)$, where $\bm{d}^{(k)}=(d_1^{(k)}, \ldots, d_{q^{(k)}}^{(k)})$ $\in\mathbb{N}_+^{q^{(k)}}$,  $\bm{t}^{(k)}= (t_0^{(k)}, \ldots, t_{q^{(k)}}^{(k)})\in\mathbb{N}_+^{q^{(k)}+1}$ with $t_u^{(k)}\le d_u^{(k)}$ for $u=0,\ldots, q^{(k)}$, 
$\bm{\beta}^{(k)} := (\beta_0, \ldots, \beta_{q^{(k)}}^{(k)})\in\mathbb{R}_+^{q^{(k)}+1}$. Note that this density class $\mathcal{H}$  includes many popular models
studied in literature, both Gaussian and non-Gaussian; see Section \ref{SEC:Examples}.

Throughout the paper, we explore $\pi_k$ in some complicated $\mathcal{G}$ with group-specific parameters $q^{(k)}, \bm{d}^{(k)}, \bm{t}^{(k)}$ and $\bm{\beta}^{(k)}$.  The selection range of the truncation parameter $J$ is based on the asymptotic order provided in Assumptions \ref{A4} and \ref{A5} in the next section.
Although $\pi^{(J)}_k$ has $J$ arguments, it involves at most  $t_0^{(k)}d_1^{(k)}$ effective arguments, implying that the two population densities differ by a small number of variables. Relevant conditions are necessary for high-dimensional classification. For instance, 
in high-dimensional Gaussian data classification, \cite{Cai:Zhang:19, Cai:Zhang:19b} show that, to consistently estimate Bayes classifier, it is necessary 
that the mean vectors differ at a small number of components.
The modular structure holds for arbitrary $J$, which may be viewed as a extension of \cite{Schmidt:19} in the functional data analysis setting.

\subsection{Convergence of Kullback-Leibler divergence}


\subsubsection{Kullback-Leibler divergence}

For the true probability distribution $\bm \pi(\bm x) = \left(\pi_1(\bm x), \ldots, \pi_{K}(\bm x)\right)^\intercal$ and any generic estimation $\widehat{\bm \pi}(\bm x) = \left(\widehat{\pi}_{1}(\bm x), \ldots, \widehat{\pi}_{K}(\bm x)\right)^\intercal$, define the corresponding discrete version Kullback-Leibler divergence 
$$\textit{KL}\left({\bm \pi}(\bm x), \widehat{\bm \pi}(\bm x) \right) = \sum_{k=1}^K \pi_{k}(\bm x) \log\left( \frac{\pi_{k}(\bm x)}{\widehat{\pi}_{k}(\bm x)}\right).$$ 

For any $\widehat{\bm \pi}$ trained with $\left\{ \left(X_i(s), \bm Y_i\right)\right\}_{i=1}^n$, we evaluate its performance by the log-likelihood ratio
$$\mathbb{E}\left[ \sum_{k=1}^K Y_k \log\left( \frac{\pi_{k}(\bm \xi)}{\widehat{\pi}_{k}(\bm \xi)}\right)\right] =   \mathbb{E} \left[\textit{KL}\left({\bm \pi}(\bm \xi), \widehat{\bm \pi}(\bm \xi) \right)\right].$$
Note that the estimation risk is associate with the well-known CE loss in Section \ref{sec:classifier}. 

Different from  the popular least  square loss,  the CE loss is the expectation with respect to the input distribution
of the Kullback-Leibler divergence of the conditional class probabilities. If anyone of the conditional class probabilities has zero estimation while the
underlying conditional class probability is positive, the risk can even become
infinite. To avoid the infinite risk, we   truncate the CE loss function and derive convergence rates without assuming either the true  conditional class probabilities or the estimators away from zero. Instead,
our misclassification risks depend on an index quantifying the behaviour of the conditional class
probabilities near zero.

Given an absolute constant $C_0\geq 2$, for any classifier $\widehat {\bm \pi} $, define the truncated Kullback-Leibler risk for some density $\bm h$ as 
\begin{equation}
    R_{\bm h,C_0}\left(\widehat{\bm \pi}\right) = {\E}_{\bm h}\left\{\sum_{k=1}^K \pi_{k}(\bm \xi)\left[C_0\wedge \log\left(\frac{ \pi_{k}(\bm \xi)}{\widehat {\pi}_k(\bm \xi)} \right) \right]\right\}.
\end{equation}

\begin{remark}
$C_0$ is commonly introduced to avoid infinity value of the ordinary Kullback-Leibler risk. It is trivial to show that $C_0$ can only be abandoned when $\widehat{\pi}_k$ is lower bounded by $\exp{(-C_0)}$ for all $k$. When $\widehat{\pi}_k $ has a large deviation from $\pi_k$ for some $k$, the Kullback-Leibler risk explodes to infinity, which makes it is infeasible to evaluate the performance of $\widehat{\bm\pi}$.

\end{remark}

\subsubsection{Convergence rate for fully observed functional data}

In this section, we provide the non-asymptotic Kullback-Leibler risk of  mfDNN classifier. 
Let 
\begin{equation}\label{EQ:ku}
    (\widehat{k}, \widehat{u}) = \arg\max_{\substack{(k,u), k=1\ldots,K, \\u=0, \ldots, q^{(k)}}}  \frac{t_u^{(k)}}{{\tilde{\beta}_u^{(k)}}}, 
\end{equation}
where ${\tilde{\beta}_u^{(k)}}=\beta_u^{(k)}\prod_{\ell=u+1}^{q^{(k)}}\beta_\ell^{(k)}\wedge 1$, and 
\begin{equation*}
    (\widehat{l}, \widehat{v}) = \arg\max_{\substack{(k,u), k=1\ldots,K, \\u=0, \ldots, q^{(k)}}}  \frac{t_u^{(k)}}{ {\tilde{\beta}_u^{(k)}}(1+\alpha_k) }.
\end{equation*}
Let $\theta=\frac{(1+\alpha_{\widehat{l}})\tilde{\beta}_{\widehat{v}}^{(\widehat{l})}}{(1+\alpha_{\widehat{l}})\tilde{\beta}_{\widehat{v}}^{(\widehat{l})} + t_{\widehat{v}}^{(\widehat{l})}}$  and $\nu=  \frac{\theta t_{\widehat{u}}^{(\widehat{k})}}{{\tilde{\beta}_{\widehat{u}}^{(\widehat{k})}} \left( 1+\tilde{\alpha}\right)}$, where $\widetilde{\alpha}  = \min_k \alpha_k\wedge 1$.

\begin{Assumption}\label{A4}
There exist some constants $C_1, C'_2, C_2, C_3$, only depending on $\mathcal{H}$ and $C_0$, such that the DNN class $\mathcal{F}(L,J,\bm{p},s)$ satisfies 
\begin{enumerate}[label=(\alph*)]
    \item\label{A4:a} $L \leq C_1\log n$; 
    \item\label{A4:b} $C'_2n^{\theta/\rho} \leq J \leq C_2 n^\nu$; 
    \item\label{A4:c} $ \max_{1\leq \ell \leq L} p_\ell \leq  C_2 n^\nu $;
    \item\label{A4:d} $s\leq  C_3 n^\nu \log n$.
\end{enumerate}
\end{Assumption}
Assumption \ref{A4} provides exact orders on $(L,\bm{p}, s)$ for network, respectively. Assumption \ref{A4}\ref{A4:b} provides the precise range on $J$. It is worth mentioning this condition implies $\varrho\ge \theta \nu^{-1}$, i.e., the function $\zeta(J)$ converges to zero in a relatively fast rate when $J\to\infty$.

In the following, we provide the convergence rate in the ideal case when the entire functional curve is fully observed.

\begin{theorem}\label{thm:rate}
There exist a positive constant $\omega_1$, only depending on $\mathcal{H}$ and $C_0$, such that 
$$\sup_{\bm h\in \mathcal{H}} R_{\bm h, C_0}\left(\widehat{\bm \pi}\right) \le \omega_1 n^{-\theta}\log^3 n,$$
where network classifier $\widehat{\bm \pi}$ belongs to $\mathcal{F}(L,J,\bm{p},s)$ in Assumption \ref{A4}.
\end{theorem}

\begin{remark}
Theorem \ref{thm:rate} provides the upper bound of the KL misclassification risk of the proposed mfDNN. When $K=2$, the multiclass classification downgrades to the binary classification problem. Compared with the minimax excess misclassification risk derived in \cite{wang:etal:22} for the binary classification, the upper bound rate in Theorem \ref{thm:rate} provides a slightly larger order, Specifically, the leading term  of the upper bound rate in Theorem \ref{thm:rate} is  $n^{-\theta}$ with $\theta=\frac{(1+\alpha )\beta} {(1+\alpha) {\beta}  + t }$, while the leading terms in Theorem 1 in \cite{wang:etal:22} is $n^{-S_0}$ with $S_0= \frac{(\alpha+1)\beta }{(\alpha+2)\beta  + t}$. Hence, when $\beta$'s, i.e. the degrees of H\"{o}lder smooth functions in the class $\mathcal{H}$, are large enough, the discrepancy between these two bounds are rarely negligible.  
Hence, to reduce potential slightly larger   risks, we recommend \cite{wang:etal:22} for the binary classification problems and when the   distribution functions are not smooth enough. Meanwhile, we recommend the mfDNN classifier  for multiclassification problems  regardless the smoothness of the   conditional distribution functions.

\end{remark}

\subsubsection{Convergence  rate for discretely observed functional data}
Practically, it is usually unrealistic to observe the full trajectory of each individual, thus the rate in Theorem \ref{thm:rate} can only be reached if sampling frequency is dense enough. Hence, it is interesting to  discuss the upper bound of the risk of mfDNN classifier when functional data are discretely observed at $m$ occasions for each subject.
Let $\widetilde{\beta} =  \max_{k=1\ldots,K}\left({\widetilde{\beta}_0^{(k)}} \wedge 1\right)$, $\theta'=\tau\widetilde{\beta}$, and $\nu'= \frac{\theta't_{\widehat{u}}^{(\widehat{k})}}{{\widetilde{\beta}_{\widehat{u}}^{(\widehat{k})}}\left(1+\widetilde{\alpha}\right)}$, where $\widehat{k}$ is defined in (\ref{EQ:ku}) and $\tau$ is a positive universal constant. 
\begin{Assumption}\label{A5}
There exist some constants ${C}_1, {C}'_2, {C}_2, C_3$, $\widetilde{C}_1, \widetilde{C}'_2, \widetilde{C}_2, $ and $\widetilde{C}_3$ only depending on $\mathcal{H}$, $C_0$ and $\tau$, and a phase transition point $m^\ast\in \mathbb{N}^+$, such that the DNN class $\mathcal{F}(L,J,\bm{p},s)$ satisfies 
\begin{enumerate}[label=(\alph*)]
    \item\label{A5:a} $L \leq {C}_1\log n \mathbb{I}(m\geq m^\ast) + \widetilde{C}_1\log m \mathbb{I}(m< m^\ast)$; 
    \item\label{A5:b} ${C}'_2n^{\theta/\rho}\mathbb{I}(m\geq m^\ast)  +  \widetilde{C}'_2 m^{\theta'/\rho}\mathbb{I}(m< m^\ast)\leq J \leq C_2 n^\nu \mathbb{I}(m\geq m^\ast) + \widetilde{C}_2 m^{\nu'} \mathbb{I}(m< m^\ast)$; 
    \item\label{A5:c} $ \max_{1\leq \ell \leq L} p_\ell \leq  {C}_2 n^\nu \mathbb{I}(m\geq m^\ast) +  \widetilde{C}_2 m^{\nu'}\mathbb{I}(m< m^\ast)$;
    \item\label{A5:d} $s\leq  {C}_3 n^\nu \log n \mathbb{I}(m\geq m^\ast) + \widetilde{C}_3m^{\nu'} \log m \mathbb{I}(m< m^\ast) $.
\end{enumerate}
\end{Assumption}
Similar to Assumption \ref{A4}, Assumption \ref{A5} provides exact orders on $L,\bm{p}, s$, and range of $J$ when sampling frequency $m$ is involved. When $m\geq m^\ast$, Assumption \ref{A5} coincides with Assumption \ref{A4} for dense functional data.

The following theorem provides the phase transition rate when functional data are discretely observed on $m$ locations at a certain rate with respect to $\tau$.
\begin{theorem}\label{thm:discrete}
When $E|\xi_j - \widehat{\xi}_j| \lesssim m^{-\tau}$ for all $j=1,\ldots, J$, there exists positive constants $\omega_1, \omega_2$ and $\omega_3$ only depending on $\mathcal{H}$, $C_0$ and $\tau$, such that 
$$\sup_{\bm h\in \mathcal{H}} R_{\bm h, C_0}\left(\widehat{\bm \pi}\right) \le \omega_1 n^{-\theta}\log^3 n\mathbb{I}(m\geq m^\ast) +\omega_2 m^{-\theta'}\mathbb{I}(m< m^\ast),$$
where $m^\ast = \lfloor \left(\omega_3 n^{\theta}/\log^{3} n\right)^{1/\theta'} \rfloor$, and $\widehat{\bm \pi}\in \mathcal{F}(L,J,\bm{p},s)$ defined in Assumption \ref{A5}.
\end{theorem}

\begin{remark}
When functional curves are discretely observed, Theorem \ref{thm:discrete} provides the convergence rate of truncated KL risk when the biases of projection scores are uniformly bounded by $m^{-\tau}$. This assumption does not hold universally for all scenarios. Nevertheless, it can be satisfied in various examples. 

For any empirical process  on $\left[ 0, 1\right]$, if we use Fourier basis with $m$ terms to decompose the curve, we can show that $E|\xi_j - \widehat{\xi}_j| \leq O\left(\max_{k=1,\ldots, K}\sqrt{\sum_{j=m+1}^\infty\left( \lambda_{kj} + \mu^2_{kj}\right)}\right)$, where $\mu_{kj} = \E\xi_j$ in the $k$-th group. As $\left\{ \lambda_{kj}\right\}_{j=1}^{\infty}$ and  $\left\{  \mu^2_{kj}\right\}_{j=1}^{\infty}$ are both convergent, when $\lambda_{kj}$ and $\mu^2_{kj}$ are decreasing no faster than some polynomial order, the assumption easily follows.

Another well-known example is the FPCA provided by {\cite{Hall:Hosseini-Nasab:06}}. According  to  Theorem 1 in \cite{Hall:Hosseini-Nasab:06}, for all $j=1,\ldots, J$, the estimators of projection scores satisfy $\E |\xi_j - \widehat{\xi}_j | \leq \max_{k}\left(\int_{\left[ 0, 1\right]^d\times \left[ 0, 1\right]^d}\left( \Omega_k  - \widehat{\Omega}_k\right)^2(s,s')dsds' \right)^{1/2}$. Hence,    $E|\xi_j - \widehat{\xi}_j| \lesssim m^{-\tau}$ holds easily when $\max_{k}\left(\int_{\left[ 0, 1\right]^d\times \left[ 0, 1\right]^d}\left( \Omega_k  - \widehat{\Omega}_k\right)^2(s,s')dsds' \right)$ is bounded properly.  
\end{remark}

\section{Examples}\label{SEC:Examples}
In this section, we provide two examples of exponential families to
justify the validation of our model assumptions, and emphasize the necessity of applying DNN approach owing to the complicated structure of data population. For simplicity, we assume the prior probability satisfies $\mathbb{P}(\bm Y=\bm e_k) = k^{-1}$ for all $k$ throughout the section.

\subsection{Independent exponential family}\label{sec:example 1}
We first consider independent projection scores, which are from exponential families. For some collection of unknown parameters $ \left\{ \theta_{kj}\right\}_{k=1,j=1}^{K,\infty} $, and unknown collections of functions $\left\{ \eta_{kj}\right\}_{k=1,j=1}^{K,\infty} $, $\left\{ U_{kj}\right\}_{k=1,j=1}^{K,\infty} $, and $\left\{ W_{kj}\right\}_{k=1,j=1}^{K,\infty} $,
we consider the $k$-th class conditional density, such that 
$$\bm\xi|\left\{\theta_{kj}\right\}_{j=1}^\infty, \bm Y=\bm e_k\sim h_k(\bm x) = \exp\left( \sum_{j=1}^\infty \eta_{kj}(\theta_{kj})U_{kj}(x_j) + W_{kj}(x_j) \right).$$
 For all $1\leq k,k'\leq K$, let {$A_{kk'} = \left\{ j: \eta_{kj}(\theta_{kj})U_{kj}(x_j)\neq\eta_{k'j}(\theta_{k'j})U_{k'j}(x_j), \forall x_j\in\mathbb{R} \right\}$ } and $B_{kk'} = \left\{ j: W_{kj}\neq W_{k'j} \right\}$ be the two sets identifying the difference between $h_k$ and $h_{k'}$. Therefore, we have the pairwise log likelihood
\begin{eqnarray*}
\log\left(h_k/h_{k'}\right) &=& \sum_{j\in \left(A_{kk'}\bigcup B_{kk'}\right)} \left\{\left[ \eta_{kj}(\theta_{kj})U_{kj}(x_j) - \eta_{k'j}(\theta_{k'j})U_{k'j}(x_j)\right] \right.\\
&&+ \left.\left[ W_{kj}(x_j) - W_{k'j}(x_j)\right]\right\}.
\end{eqnarray*}
Given some universal  constant $N_{kk'}$, when $|A_{kk'}\bigcup B_{kk'}|\leq N_{kk'}$, there exists a positive integer $J_{max} = \max  \bigcup_{k,k'} \left( A_{kk'}\bigcup B_{kk'} \right)$, such that $\pi_{k}=\pi_{k}^{(J)}$ for all $J\geq J_{max}$. 

By definition, Assumption \ref{A1} holds for $\alpha=1$. Assumption \ref{A3} holds when $h_k/h_{k'}$ is bounded for all pairs. Note that it is trivial when $\{h_k\}_{k=1}^K$ share the same $ \left\{ U_{kj}\right\}_{k=1,j=1}^{K,\infty}$  and $\left\{ W_{kj}\right\}_{k=1,j=1}^{K,\infty} $ , such as Gaussian distribution, student's t distribution and exponential distribution, whose density ratio of their kind is always bounded. Assumption \ref{A2} holds for $J_0 = J_{max}$, and for arbitrary function $e(\cdot)$ with exponential tails and density $\pi(\cdot)$. Since $\pi_k = \left[1+\sum_{k'\neq k} \log(h_k/h_{k'})\right]^{-1}$, the smoothness is determined by $\left\{ \eta_{kj}\right\}_{k=1,j=1}^{K,\infty} $, $\left\{ U_{kj}\right\}_{k=1,j=1}^{K,\infty} $, and $\left\{ W_{kj}\right\}_{k=1,j=1}^{K,\infty} $, thus  $\bm h$ is trivially in some $\mathcal{H}$.

\subsection{Exponential family with in-block interaction}
In this example, we consider $\xi_j$ are dependent with each other in a block, but independent across blocks, which is an extension of the example in Section \ref{sec:example 1}. Given a sequence of positive integers $\{\ell_p\}_{j=1}^\infty$, such that $0=\ell_1<\ell_2<\ldots$, we define {the $p$-th group index set $\mathcal{E}_p = \{ \ell_{p} + 1, \ldots, \ell_{p+1}\}$, such that the cardinality $|\mathcal{E}_p| = \ell_{p+1} - \ell_{p}$,} therein grouping are based on adjacent members for simplicity. 
For a collection of unknown parameters $ \left\{ \bs\theta_{kj}\right\}_{k=1,p=1}^{K,\infty}$, 
and unknown collections of functions $\left\{ \widetilde{\eta}_{kp}\right\}_{k=1,p=1}^{K,\infty} $, $\left\{  \widetilde{U}_{kp}\right\}_{k=1,p=1}^{K,\infty} $, and $\left\{  \widetilde{W}_{kp}\right\}_{k=1,p=1}^{K,\infty} $, where $\widetilde U_{kp}$ and $\widetilde W_{kp}$ are functions from $\mathbb{R}^{|\mathcal{E}_p|}$ to $\mathbb{R}$.


  Consider the joint conditional density $$h_k(\bm x) = \exp\left( \sum_{p=1}^\infty  \eta_{kp}(\bm\theta_{kp})\widetilde U_{kp} (\bs x_p)   + \widetilde W_{kp} (\bs x_p) \right)$$
for class $k$, where $\bs x_p = \left(x_j\right)_{j=\ell_p+1,\ldots,\ell_{p+1}}$.  

For any $1\leq k,k' \leq K$, define the density difference sets 
$$\widetilde{A}_{kk'} = \left\{ p: \widetilde\eta_{kp}(\bm\theta_{kp})\widetilde U_{kp}(\bs x_p)\neq\eta_{k'p}(\bm\theta_{k'p})\widetilde U_{k'p}(\bs x_p), \forall \bs x_p\in\mathbb
{R}^{|\mathcal{E}_p|}\right\}$$
and $\widetilde{B}_{kk'} = \left\{ p: \widetilde{W}_{kp}\neq \widetilde{W}_{k'p} \right\}$, and the pairwise log likelihood is thus given by  
 \begin{eqnarray*}
&&\log\left(h_k/h_{k'}\right)\\
&=& \sum_{p\in \widetilde{A}_{kk'}\bigcup \widetilde{B}_{kk'}} \left\{\left[ \widetilde\eta_{kp}(\bm\theta_{kp})\widetilde{U}_{kp} (\bs x)  - \widetilde\eta_{k'p}(\bm\theta_{k'p})\widetilde U_{k'p}(\bs x) \right]+\left[\widetilde W_{kp}(\bs x) - \widetilde W_{k'p}(\bs x)\right]\right\}.
\end{eqnarray*} 
Given some finite positive number $N_{kk'}$, such that $|A_{kk'}\bigcup B_{kk'}|\leq N_{kk'}$, the  verification can be similarly derived from   Section \ref{sec:example 1}.

\section{Simulation studies}\label{SEC:simulation}
In this section, we provide  numerical evidences to demonstrate the superior performance of mfDNN.  In  all simulations, we generated $n_k=200,350,700$ training samples for each class, and testing sample sizes $100, 150, 300$, respectively. Based on the fact that there is no existing multi-classification method specifically designed for multidimensional functional data,  for comparison, we include the sparse discriminate analysis and $\ell_1$ penalized Fisher’s discriminant analysis (MSDA) approach introduced in \cite{Mai:19} and penalized  linear discriminant analysis (PLDA) classifier in \cite{Witten:Tibshirani:11}. In fact, MSDA and PLDA are   efficient classifiers designed for high-dimensional i.i.d. observations. 
To make these two methods directly applicable to functional data, we first pre-processed 2D or 3D functional data by vectorization. The realization is via the \texttt{R} packages \texttt{msda} and \texttt{PenalizedLDA}, where the tuning parameter candidates for the penalty term are generated by default.
We use the default five-fold and six-fold cross-validation to tune MSDA and PLDA, respectively. For mfDNN, we use tensor of Fourier basis to extract projection scores by integration. The structure parameters $(L,J, \bm p, s)$ are selected by Algorithm \ref{alg:mfdnn}, where the candidates are given based on Theorem \ref{thm:discrete}.
We summarize R codes and examples for the proposed mfDNN algorithms   on \texttt{GitHub}  (\url{https://github.com/FDASTATAUBURN/mfdnn}).

\subsection{2D functional data}\label{sec:sim_2d}
For $k=1,2,3$, we generated functional data $X_{i}^{(k)}(s,s')= \sum_{j=1}^5 \xi_{ij}^{(k)} \psi_j(s,s')$, $s,s'\in \left[ 0, 1\right]$. Let $\psi_1(s,s')=s$, $\psi_2(s,s')=s'$ , $\psi_3(s,s')=ss'$, $\psi_4(s,s')=s^2s'$, $\psi_5(s,s')=s{s'}^2$. Define $\mathbf{1}_k$ be a $k\times1$ vector with all the elements one. We specify the distribution of $\xi_{ij}^{(k)}$'s as following.

\textit{Model 1} (2D Gaussian):  Let $\left(\xi_{i1}^{(k)}, \ldots, \xi_{i5}^{(k)}\right)^\intercal \sim N(\bm\mu_k, \bm\Sigma_k)$, where  $\bm\mu_1= (4,4,3,3,3)^\intercal$, $\bm\Sigma_1^{1/2}= \text{diag}\left(8,7,6,5,4\right) $,   $\bm\mu_2= -\mathbf{1}_5$,  $\bm\Sigma_2^{1/2}=\text{diag}\left( 5,4,3,2,1\right) $, $\bm\mu_3=  \mathbf{0}_5$,  $\bm\Sigma_3^{1/2}= \text{diag}\left(2.5, 2, 1.5, 1, 0.5\right) $.

\textit{Model 2} (2D Mixed 1):  
Let $\left(\xi_{i1}^{(k)}, \ldots, \xi_{i5}^{(k)}\right)^\intercal \sim N(\bm\mu_k, \bm\Sigma_k)$ for $k=1,2$, and $\xi_{ij}^{(3)} \sim t_{2j+1}(\nu_{j})$,  where  $\bm\mu_1= -\mathbf{1}_5$, $\bm\Sigma_1^{1/2}= \text{diag}\left( 5,4,3,2,1\right) $,   $\bm\mu_2= \mathbf{0}_5$,  $\bm\Sigma_2^{1/2}=\text{diag}\left( \frac{5}{2}, 2, \frac{3}{2}, 1, \frac{1}{2}\right) $, $(\nu_{1},\ldots,\nu_{5})^\intercal= 3\cdot\mathbf{1}_5$.

\textit{Model 3} (2D Mixed 2):  Let $\left(\xi_{i1}^{(1)}, \ldots, \xi_{i5}^{(1)}\right)^\intercal \sim N(\bm\mu_k, \bm\Sigma_k)$, $\xi_{ij}^{(2)} \sim t_{j+1}(\nu_{2j})$, $\xi_{ij}^{(2)} \sim t_{2j+1}(\nu_{3j})$,  where  $\bm\mu_1= \mathbf{0}_5$, $\bm\Sigma_1^{1/2}= \text{diag}\left( \frac{5}{2}, 2, \frac{3}{2}, 1, \frac{1}{2}\right)$,  $(\nu_{21},\ldots,\nu_{25})^\intercal= \mathbf{1}_5$, $(\nu_{31},\ldots,\nu_{35})^\intercal= 3\cdot\mathbf{1}_5$.

\textit{Model 4} (2D Mixed 3):  Let  $\xi_{ij}^{(1)} \sim \textit{Exp}(r_{1j})$,  $\xi_{ij}^{(2)} \sim t_{2j+1}(\nu_{2j})$, and $\left(\xi_{i1}^{(3)}, \ldots, \xi_{i5}^{(3)}\right)^\intercal$ $ \sim N(\bm\mu_3, \bm\Sigma_3)$, where $(r_{11}, \ldots, r_{15})^\intercal = (0.1,0.3,0.5,0.7,0.9 )^\intercal$, $(\nu_{21},\ldots,\nu_{25})^\intercal= 3\times\mathbf{1}_5$, $\bm\mu_3= \mathbf{0}_5$, $\bm\Sigma_3^{1/2}= \text{diag}\left( 2.5, 2, 1.5, 1, 0.5\right) $.

For each model, we observe the functional data on $3\times 3$, $5\times 5$, $10\times 10$, and $20\times 20$ grid points over $\left[ 0, 1\right]^2$, respectively. As a result, the sampling frequency $m=9,25,100,400$, which indicates that the functional observations are from sparse to dense. Tables \ref{TAB:sim_2d_1} and \ref{TAB:sim_2d_2} demonstrate  the results of $100$ simulations. For mfDNN, it can be seen that the misclassification risks decrease as  the sample size $n$ increasing, as well as the increase of the sampling frequency $m$. This founding further confirms  Theorem \ref{thm:discrete}. Given the relatively sparse sampling frequency, i.e., $m=9$,  MSDA  has slightly better performance than mfDNN does. However, despite the increase of $m$ in Table \ref{TAB:sim_2d_2},    there is no improvement of both MSDA and PLDA  methods in terms of misclassification risks. This finding indicates that MSDA and PLDA classifiers can not be improved with more gathered information. In summary, the simulation results illustrate that  the proposed mfDNN method outperforms the existing sparse   and penalized discriminant analysis when classifying   2D dense functional data.

\subsection{3D functional data}\label{sec:sim_3d}
For $k=1,2, 3$ and $s_1,s_2,s_3\in \left[ 0, 1\right]$, we generate 3D functional data $X_{i}^{(k)}(s_1,s_2,s_3)= \sum_{j=1}^{9} \xi_{ij}^{(k)} \psi_j(s_1,s_2,s_3)$ where $\psi_1(s_1,s_2,s_3)=s_1$, $\psi_2(s_1,s_2,s_3)=s_2$,
$\psi_3(s_1,s_2,s_3)=s_3$,
$\psi_4(s_1,s_2,s_3)=s_1s_2$,
$\psi_5(s_1,s_2,s_3)=s_1s_3$,
$\psi_6(s_1,s_2,s_3)=s_2s_3$,
$\psi_7(s_1,s_2,s_3)=s_1^2$,
$\psi_8(s_1,s_2,s_3)=s_2^2$,
$\psi_9(s_1,s_2,s_3)$ $=s_3^2$, and the distribution of $\xi_{ij}^{(k)}$'s are specified as below.

\textit{Model 5} (3D Gaussian):  Let  $\left(\xi_{i1}^{(k)}, \ldots, \xi_{i9}^{(k)}\right)^\intercal \sim N(\bm\mu_k, \bm\Sigma_k)$, where  $\bm\mu_1= 2\times\mathbf{1}_9$, $\bm\Sigma_1^{1/2}= \bm\Sigma_2^{1/2}=\text{diag}\left( 9,8,7,6,5,4, 3, 2, 1\right) $,   $\bm\mu_2=\bm\mu_3=  \mathbf{0}_9$, $\bm\Sigma_3^{1/2}= 1/3\times\bm\Sigma_1^{1/2} $.

\textit{Model 6} (3D Mixed 1):  Let $\left(\xi_{i1}^{(k)}, \ldots, \xi_{i9}^{(k)}\right)^\intercal \sim N(\bm\mu_k, \bm\Sigma_k)$ for $k=1,2$, and $\xi_{ij}^{(3)} \sim t_{j+1}(\nu_{j})$,  where  $(\nu_{1},\ldots,\nu_{5})^\intercal= 3\times\mathbf{1}_5$, $\bm\mu_1= -\mathbf{1}_9$, $\bm\Sigma_1^{1/2}= \text{diag} ( 5.5, 5, 4.5, 4, $ $3.5, 3, 2.5, 2, 1.5) $,   $\bm\mu_2= \mathbf{0}_9$,  $\bm\Sigma_2^{1/2}=\text{diag}\left( 4.5, 4, 3.5, 3, 2.5, 2, 1.5, 1, 0.5\right)$.

\textit{Model 7} (3D Mixed 2):   Let $\left(\xi_{i1}^{(1)}, \ldots, \xi_{i9}^{(1)}\right)^\intercal \sim N(\bm\mu_k, \bm\Sigma_k)$, $\xi_{ij}^{(2)} \sim t_{j+1}(\nu_{2j})$, where  $\bm\mu_1= \mathbf{0}_9$, $\bm\Sigma_1^{1/2}= \text{diag}\left(4.5, 4, 3.5, 3, 2.5, 2, 1.5, 1, 0.5\right)$,  $(\nu_{21},\ldots,\nu_{29})^\intercal= -\mathbf{1}_9$, $(\nu_{31},\ldots,\nu_{39})^\intercal= 0.5\times\mathbf{1}_9$.

\textit{Model 8} (3D Mixed 3):  Let $\xi_{ij}^{(1)} \sim \textit{Exp}(r_{1j})$,  $\xi_{ij}^{(2)}$ $ \sim t_{j+1}(\nu_{2j})$, and $\left(\xi_{i1}^{(3)}, \ldots, \xi_{i9}^{(3)}\right)^\intercal$ $ \sim N(\bm\mu_3, \bm\Sigma_3)$, where $(r_{11}, \ldots, r_{19})^\intercal = 0.1\times(1,3,5,7,9,11,13,15,17 )^\intercal$, $(\nu_{21},\ldots,\nu_{29})^\intercal$ $= 0.6\times\mathbf{1}_9$, $\bm\mu_3= \mathbf{0}_9$, $\bm\Sigma_3^{1/2}= \text{diag}\left( 4.5, 4, 3.5, 3, 2.5, 2, 1.5, 1, 0.5\right) $.

For the 3D functional data, we apply similar setups as 2D cases. We observe the functional data on $2\times 2\times 2$, $3\times 3\times 3$, $4\times 4\times 4$, and $5\times 5\times 5$ grid points over $\left[ 0, 1\right]^3$, respectively, and the sampling frequency $m=8,27,64,125$. Tables \ref{TAB:sim_3d_1} and \ref{TAB:sim_3d_2} demonstrate the results of $100$ simulations. The proposed mfDNN classifier is superior to its counterparts for all 3D functional data cases. Meanwhile, there also exists the phase transition patterns for mfDNN method.  However, the performance of MSDA and PLDA methods lacks of improvement with the increase of $m$.  It can be seen that when $m=125$, the misclassification error rates of mfDNN are almost one third of MSDA's and one forth of PLDA's in Gaussian case, and almost a half of either MSDA's or PLDA's error rates in  Models 7 and 8. A plausible reason is that given the functional data framework, our proposed mfDNN can properly accommodate the repeatedly observed  data over pixels or voxels, while other competitors only treat those information as common high-dimensional covariates and ignore the underlining smoothing structures. By efficiently extracting the projection scores of the continuum, the proposed mfDNN has full potential to discover the underlying distributions of the functional data clusters.    This again  demonstrates   our proposed classifier has a distinct advantage over these competitors in complex imaging data classification problems.

\begin{table}
\caption{\label{TAB:sim_2d_1} Averaged misclassification rates with standard errors in brackets for 2D simulations when $m=9$ and $m=25$  over $100$
replicates.}
\centering
\begin{tabular}{@{\extracolsep{0.1pt}} ccccccccc}
\hline
\hline   
\multirow{2}{*}{Model} & \multirow{2}{*}{$n_k$} & \multicolumn{3}{c}{$m=9$} & & \multicolumn{3}{c}{$m=25$}  \\   \cline{3-5} \cline{7-9}
 &  & \multicolumn{1}{c}{mfDNN} &  \multicolumn{1}{c}{MSDA} & \multicolumn{1}{c}{PLDA} & &\multicolumn{1}{c}{mfDNN} &  \multicolumn{1}{c}{MSDA} &\multicolumn{1}{c}{PLDA}  \\ \hline
\multirow{6}{*}{2D Gaussian} &200 & 0.259  & \textbf{0.246}  &0.323 & & \textbf{0.205}  & 0.245 &0.325   \\
  &  &  (0.031)  &  (0.027) & (0.028)  & &  (0.026)  &  (0.028) & (0.028)  \\  \cline{3-5} \cline{7-9}
 &350 & \textbf{0.232}  & 0.243  &0.325   & &  \textbf{0.200}  & 0.243 & 0.328  \\
&  &  (0.020) & (0.022) & (0.022)  & &   (0.020) &  (0.021)&  (0.022)  \\  \cline{3-5} \cline{7-9}
&700 & \textbf{0.227} & 0.241  & 0.323  & &  \textbf{0.196}   & 0.241  & 0.325 \\
& &  (0.014) &  (0.014) &(0.016)  & &  (0.014)  &  (0.015) &  (0.016)\\
\hline
\multirow{6}{*}{2D Mixed 1 } &200 & 0.158 & \textbf{0.150}  &0.227  & & \textbf{0.100}   & 0.150 & 0.229 \\
& & (0.020) &  (0.021) & (0.026)  & &  (0.018)  &  (0.022)&  (0.025) \\  \cline{3-5} \cline{7-9}
&350& 0.153  &\textbf{0.144}  &0.227    & & \textbf{0.085}  & 0.144  &0.229  \\
  & &  (0.015) & (0.016) & (0.020)  & &  (0.012) &  (0.016) & (0.019) \\  \cline{3-5} \cline{7-9}
&700 & \textbf{0.152} & 0.156  & 0.229   & & \textbf{0.081} & 0.123 & 0.231  \\
&  &  (0.011) &  (0.014) &  (0.014)  & & (0.010) &  (0.014)&  (0.014) \\ \hline
\multirow{6}{*}{2D Mixed 2 }&200 & 0.166  & \textbf{0.152}  &0.198   & & \textbf{0.140}    &0.153 & 0.200    \\
& & (0.023) &  (0.022) & (0.025)  & &  (0.020)   & (0.022)&  (0.025)   \\  \cline{3-5} \cline{7-9}
  &350 &0.165 & \textbf{0.152} & 0.200   & &  \textbf{0.135}  &0.152  & 0.202  \\
 &  & (0.016) &  (0.016)&  (0.022)  & & (0.016)  & (0.016)&  (0.022)  \\  \cline{3-5} \cline{7-9}
&700& 0.163 & \textbf{0.148} & 0.197    & &  \textbf{0.123}   & 0.148 & 0.199  \\
& & (0.011) &  (0.012)&  (0.013)   & &   (0.010)  &  (0.012)&  (0.013)  \\ \hline
\multirow{6}{*}{2D Mixed 3}&200 & 0.168  & \textbf{0.115} & 0.264  & & 0.136   & \textbf{0.116} & 0.264  \\
&  &  (0.030) &  (0.018)& (0.023)  & &  (0.022)  &  (0.018)&  (0.024) \\  \cline{3-5} \cline{7-9}
  &350 & 0.164  & \textbf{0.115} & 0.265   & & 0.132  &\textbf{0.115}  & 0.265 \\
  &  &  (0.024) &  (0.014)& (0.021)  & &   (0.019)  & (0.014) &  (0.021) \\  \cline{3-5} \cline{7-9}
&700 & 0.156   & \textbf{0.114} & 0.264  & &  0.123  &\textbf{0.114} &0.264  \\ 
&  &  (0.014)  &  (0.011)&  (0.016)  & & (0.014)    & (0.011)& (0.016) \\ 
\hline
\hline
\end{tabular}
\end{table}

\begin{table} 
\caption{\label{TAB:sim_2d_2} Averaged misclassification rates with standard errors in brackets for 2D simulations when $m=100$ and $m=400$ over $100$
replicates.}
\centering
\begin{tabular}{@{\extracolsep{0.1pt}} ccccccccc}
\hline
\hline   
\multirow{2}{*}{Model} & \multirow{2}{*}{$n_k$} & \multicolumn{3}{c}{$m=100$} && \multicolumn{3}{c}{$m=400$}  \\  \cline{3-5} \cline{7-9}
 &  & \multicolumn{1}{c}{mfDNN} &  \multicolumn{1}{c}{MSDA} &  \multicolumn{1}{c}{PLDA}&& \multicolumn{1}{c}{mfDNN} &  \multicolumn{1}{c}{MSDA}&  \multicolumn{1}{c}{PLDA} \\ \hline
\multirow{6}{*}{2D Gaussian} &200   & \textbf{0.147}  & 0.244 &0.326 & & \textbf{0.145}   &0.245 & 0.329  \\
  &   &  (0.024) &  (0.027)& (0.029)& & (0.025)  & (0.027)&  (0.029) \\\cline{3-5} \cline{7-9}
  &350  & \textbf{0.139}   & 0.242 & 0.328 & & \textbf{0.139}  &0.242 &0.331 \\ 
  &  & (0.017)  &  (0.021)&  (0.023)&& (0.017)  & (0.021) & (0.022)\\ \cline{3-5} \cline{7-9}
&700   & \textbf{0.132}   & 0.241 & 0.327 &  & \textbf{0.131}  &0.241 & 0.329 \\
&   &  (0.011)  &  (0.014)&  (0.016) & & (0.011) & (0.014)&  (0.016) \\
 \hline
\multirow{6}{*}{2D Mixed 1} &200  & \textbf{0.090}  & 0.150 & 0.229 && \textbf{0.089}  & 0.150 &0.232 \\
&   &  (0.018) & (0.021)&  (0.025)& &  (0.017)  &  (0.022)& (0.026) \\ \cline{3-5} \cline{7-9}
  &350&  \textbf{0.076}  & 0.144 & 0.229 & & \textbf{0.075}  &0.144 &0.232  \\
  & &   (0.010) &  (0.016)&  (0.020) && (0.010)  & (0.016)& (0.019) \\ \cline{3-5} \cline{7-9}
&700   &\textbf{0.070}  & 0.142  & 0.232 & & \textbf{0.069}  &0.142 & 0.235  \\
&   & (0.009) & (0.011) & (0.014)& & (0.008)  & (0.011)&  (0.015) \\
 \hline
\multirow{6}{*}{2D Mixed 2} &200   & \textbf{0.121}   & 0.152 & 0.201 && \textbf{0.119}   &0.152 & 0.204  \\
&  &  (0.019)  &  (0.022)&  (0.025)& &  (0.018)  & (0.022)& (0.025) \\ \cline{3-5} \cline{7-9}
  &350  & \textbf{0.116} & 0.152 & 0.202 & &\textbf{0.114} &0.151 &0.206   \\
 & &  (0.014) & (0.016)&  (0.021) && (0.013) & (0.016)& (0.021)  \\ \cline{3-5} \cline{7-9}
&700  & \textbf{0.108} & 0.148 &0.199 & & \textbf{0.069}  &0.148 &0.203 \\
&  &  (0.009)&  (0.012)& (0.012) && (0.009)  & (0.012)& (0.013) \\
 \hline
\multirow{6}{*}{2D Mixed 3}&200  & \textbf{0.107}  & 0.116 &0.264 & &\textbf{0.099}   & 0.116 &0.264 \\
&   & (0.024) & (0.018)& (0.023) & &(0.020)  &  (0.019)& (0.024)\\ \cline{3-5} \cline{7-9}
  &350  & \textbf{0.098}  &0.116 & 0.265 & &\textbf{0.090}  &  0.115 &0.267 \\
  & &  (0.012) & (0.014)&  (0.021) && (0.012) &   (0.014)& (0.021)  \\ \cline{3-5} \cline{7-9}
&700   & \textbf{0.097} &0.114 &0.265 & &\textbf{0.085}   & 0.114 &0.265 \\ 
&   &  (0.012)& (0.011)& (0.016) && (0.010)  &  (0.011)& (0.016) \\
\hline
\hline
\end{tabular}
\end{table}

\begin{table}
\caption{\label{TAB:sim_3d_1} Averaged misclassification rates with standard errors in brackets for 3D simulations when $m=8$ and $m=27$ over $100$
replicates.}
\centering
\begin{tabular}{@{\extracolsep{0.1pt}} ccccccccc}
\hline
\hline   
\multirow{2}{*}{Model} & \multirow{2}{*}{$n_k$} & \multicolumn{3}{c}{$m=8$} && \multicolumn{3}{c}{$m=27$}  \\  \cline{3-5} \cline{7-9}
 &  & \multicolumn{1}{c}{mfDNN} &  \multicolumn{1}{c}{MSDA}&  \multicolumn{1}{c}{PLDA} && \multicolumn{1}{c}{mfDNN} &  \multicolumn{1}{c}{MSDA} &  \multicolumn{1}{c}{PLDA}\\ \hline
\multirow{6}{*}{ 3D Gaussian} &200 & \textbf{0.374}   & 0.465 &0.485  & &\textbf{0.234}  &0.367 & 0.491  \\
  &  &  (0.024)  &(0.037)& (0.045)  && (0.020) & (0.030) &  (0.046) \\ \cline{3-5} \cline{7-9}
&350 & \textbf{0.373}  & 0.474 & 0.495 && \textbf{0.233}   & 0.369 & 0.500 \\
 &   & (0.020) &  (0.029)&  (0.042)& & (0.020)   &  (0.023)&  (0.041) \\ \cline{3-5} \cline{7-9}
&700 & \textbf{0.363}   & 0.472  &0.488 & & \textbf{0.232}   & 0.374 & 0.493 \\
&  &  (0.014)  &  (0.025) & (0.041) & & (0.013)  &  (0.018)&  (0.041)\\
  \hline
\multirow{6}{*}{ 3D Mixed 1} &200 & \textbf{0.215}  & 0.224 & 0.238 & & \textbf{0.168}   & 0.164 & 0.242 \\
& & (0.022) & (0.024)&  (0.022) & & (0.021)  &  (0.022)&  (0.022)  \\ \cline{3-5} \cline{7-9}
 &350& \textbf{0.200}  & 0.221 & 0.236 & &\textbf{0.151} &0.156 &0.240 \\
  & &  (0.018) &  (0.019)&  (0.020) & & (0.017)  & (0.016)& (0.020)\\ \cline{3-5} \cline{7-9}
&700 &  \textbf{0.188}  &  0.220 & 0.237  &&\textbf{0.149}   & 0.152 &0.241  \\
&  &   (0.011) &  (0.014)&  (0.012) && (0.010)   &  (0.011)& (0.013) \\
 \hline
\multirow{6}{*}{ 3D Mixed 2} &200  & \textbf{0.311}  & 0.325 & 0.346 && \textbf{0.277}  & 0.294  & 0.348 \\
&  & (0.018)  & (0.025)&  (0.026)&&  (0.020)  & (0.022) &  (0.025)\\ \cline{3-5} \cline{7-9}
 &350 & \textbf{0.307} &  0.328 & 0.345&  &\textbf{0.268}    & 0.289  &  0.348   \\
  & & (0.018)& (0.021)&  (0.022)  && (0.018)   & (0.018)  &   (0.022)   \\ \cline{3-5} \cline{7-9}
&700   & \textbf{0.253}   & 0.326 & 0.347 & & \textbf{0.234} & 0.286 &  0.350  \\
&   &  (0.014)  &  (0.015)&  (0.015)& & (0.013)  &  (0.014)&   (0.015) \\
 \hline
\multirow{6}{*}{ 3D Mixed 3} &200 & \textbf{0.272}   & 0.308 & 0.315 && \textbf{0.268}   & 0.292 &0.313 \\
&  & (0.026)  &  (0.031)&  (0.030)&&  (0.026)  &  (0.029) & (0.030)\\ \cline{3-5} \cline{7-9}
  &350 & \textbf{0.268}  & 0.303 &0.309& & \textbf{0.239}  & 0.286 &0.310 \\
  &  &  (0.019) & (0.023)& (0.024) && (0.018)  &  (0.024)& (0.023) \\ \cline{3-5} \cline{7-9}
&700 & \textbf{0.253}  & 0.302 & 0.311  && \textbf{0.234}    &0.286  & 0.312 \\
 &  &  (0.014) &  (0.017)&  (0.017) &&  (0.013)   & (0.015) &  (0.017)\\
\hline
\hline
\end{tabular}
\end{table}

\begin{table}
\caption{\label{TAB:sim_3d_2}  Averaged misclassification rates with standard errors in brackets for 3D simulations when $m=64$ and $m=125$ over $100$
replicates.}
\centering
\begin{tabular}{@{\extracolsep{0.1pt}} ccccccccc}
\hline
\hline   
\multirow{2}{*}{Model} & \multirow{2}{*}{$n_k$} & \multicolumn{3}{c}{$m=64$} && \multicolumn{3}{c}{$m=125$}  \\ \cline{3-5} \cline{7-9}
 &  & \multicolumn{1}{c}{mfDNN} &  \multicolumn{1}{c}{MSDA} &  \multicolumn{1}{c}{PLDA} &&  \multicolumn{1}{c}{mfDNN} &  \multicolumn{1}{c}{MSDA} &  \multicolumn{1}{c}{PLDA}\\ \hline
\multirow{6}{*}{3D Gaussian} &200 & \textbf{0.135}  & 0.369 & 0.493  &&\textbf{0.134}   &0.369 & 0.494 \\
 &  &  (0.018) &  (0.029)&  (0.048) && (0.017)  & (0.029)&  (0.048)\\ \cline{3-5} \cline{7-9}
&350   & \textbf{0.123} & 0.369 & 0.502 &&\textbf{0.118} &0.370 & 0.501 \\
 &  &  (0.015) &  (0.022)&  (0.042) && (0.012)& (0.022)&  (0.042) \\ \cline{3-5} \cline{7-9}
&700   & \textbf{0.114}  & 0.374 & 0.495 &&\textbf{0.108} &0.374 &0.495  \\
&   & (0.010) &  (0.019)&  (0.04) && (0.009) & (0.019)& (0.042) \\
\hline
\multirow{6}{*}{3D Mixed 1}&200  & \textbf{0.130}  &0.163 & 0.243  && \textbf{0.127}  & 0.162 &0.244  \\
&  & (0.021) & (0.022)&  (0.022) &&  (0.020)  &  (0.022)& (0.023) \\ \cline{3-5} \cline{7-9}
  &350& \textbf{0.109}  &0.156 &0.241  && \textbf{0.106} &0.155 &0.241\\
  & &  (0.014) & (0.016)& (0.020) &&  (0.014)  &(0.016)& (0.020) \\ \cline{3-5} \cline{7-9}
&700   &\textbf{0.098}  & 0.153 &0.243 && \textbf{0.098}  & 0.152 &0.242 \\
&    & (0.010) & (0.011)& (0.013) & & (0.009) & (0.011)& (0.013)\\
\hline
\multirow{6}{*}{ 3D Mixed 2}&200   & \textbf{0.166}  & 0.294 &0.350  && \textbf{0.154}   & 0.295 & 0.348 \\
&   &  (0.020) &  (0.022)& (0.025) & &  (0.019)  &  (0.024)&  (0.026)  \\ \cline{3-5} \cline{7-9}
 &350   &\textbf{0.147}  & 0.289 &0.349 &&\textbf{0.135}   & 0.289 & 0.349 \\
  &   & (0.015)  &  (0.018)& (0.023) & &(0.013)  & (0.018)&  (0.023) \\ \cline{3-5} \cline{7-9}
&700&  \textbf{0.141}   &0.286 & 0.351   & &\textbf{0.126}   & 0.286 &0.351 \\
& &   (0.012)  &  (0.015)& (0.014) & & (0.011)  &  (0.014)& (0.014)\\
\hline
\multirow{6}{*}{ 3D Mixed 3}&200  & \textbf{0.184}  & 0.292 &0.312 && \textbf{0.176}    & 0.292 &0.312  \\
&  &  (0.022) &  (0.029)& (0.030) &&  (0.021)   &  (0.028)& (0.030) \\ \cline{3-5} \cline{7-9}
 &350   & \textbf{0.171} & 0.287 & 0.310   & & \textbf{0.162}   & 0.286 & 0.310 \\
  &    &  (0.016) &  (0.023)&  (0.024) & &  (0.016)  &  (0.024)&  (0.023)\\ \cline{3-5} \cline{7-9}
&700   & \textbf{0.160}  &0.286 & 0.312 & & \textbf{0.151}   & 0.285 & 0.312 \\
&  &  (0.011) & (0.015)&  (0.017)&&  (0.012)  &  (0.015)&  (0.017)\\
\hline
\hline
\end{tabular}
\end{table}

\section{Real data analysis}\label{SEC:realdata}

\subsection{Handwritten digits}\label{sec:digits}
The first benchmark data example was extracted from the MNIST database
(\url{http://yann.lecun.com/exdb/mnist/}).  This classical MNIST database contains 60,000 training images and 10,000 testing images of handwritten digits ($0,1,\ldots, 9$), and the black and white images were normalized to fit into a $28\times 28$ pixel bounding box and anti-aliased. 
We used tensor of Fourier basis for data processing. According to our numerical experience, we choose  candidates for $(L,J, \bm p, s)$, such that $L=(2,3,4)^\intercal$, $J=(300,500,800)^\intercal$, $\|\bm p\|_\infty \ = (500,1000,2000)^\intercal$, and $s=(0.01,0.1,0.5)$ for dropout rate. Here we abuse the notation of $s$, as the dropout is the technique we choose to sparsify the neural network. With the optimal parameters $L_{opt}=3$, $J_{opt}=500$,  $\|\bm p_{opt}\|_\infty \ = 1000$, $s_{opt}=0.01$ through validation, we demonstrate the misclassification risk in  Table \ref{TAB:MNIST}. We estimate the rules given by MSDA, PLDA and our proposal on the training set. As most observations for each subject are zeros, \texttt{PenalizedLDA}  reports errors and does not work any more. 
It can be seen that our proposal achieves the higher accuracy with
the sparsest classification rule. This again shows that our method is a   competitive classifier and has more broader applications. 

\begin{figure}
\caption{\label{fig:number} Samples from MNIST data.}
\begin{center}
\includegraphics[width = 0.18\textwidth]{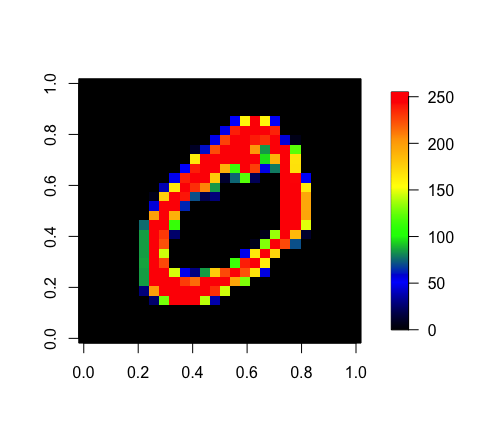} 
\includegraphics[width = 0.18\textwidth]{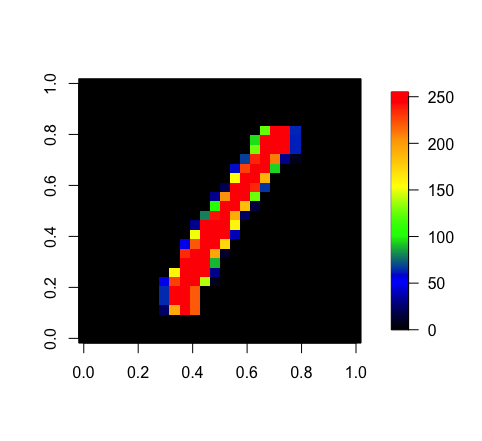} 
\includegraphics[width = 0.18\textwidth]{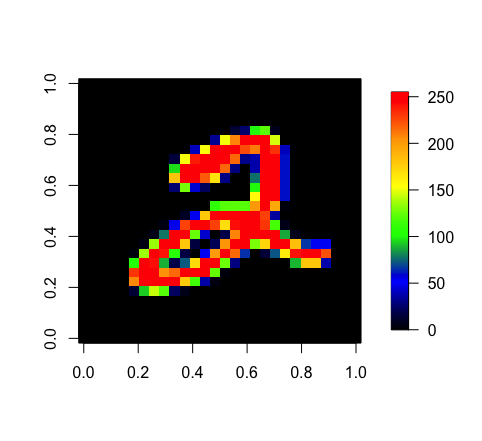}
\includegraphics[width = 0.18\textwidth]{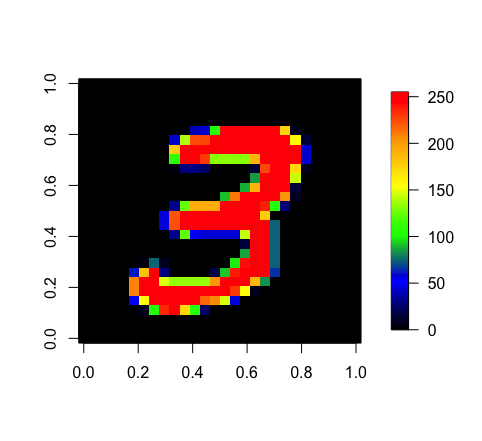} 
\includegraphics[width = 0.18\textwidth]{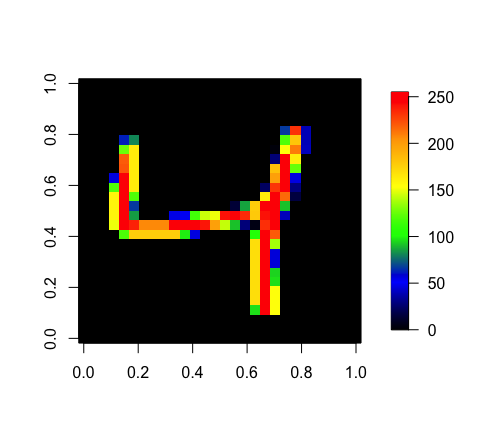}
\\
\includegraphics[width = 0.18\textwidth]{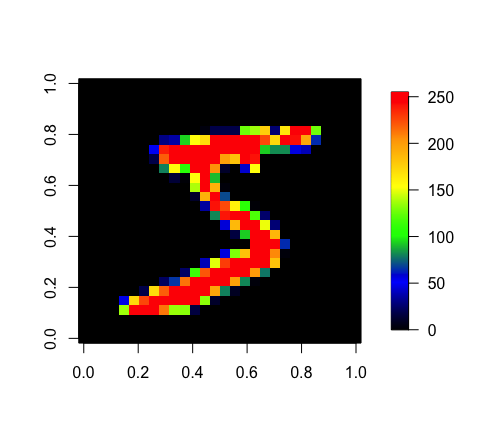} 
\includegraphics[width = 0.18\textwidth]{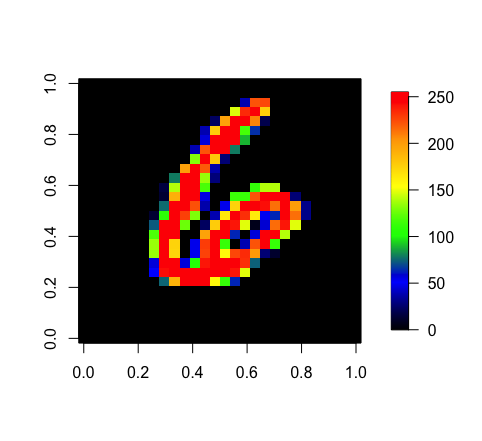} 
\includegraphics[width = 0.18\textwidth]{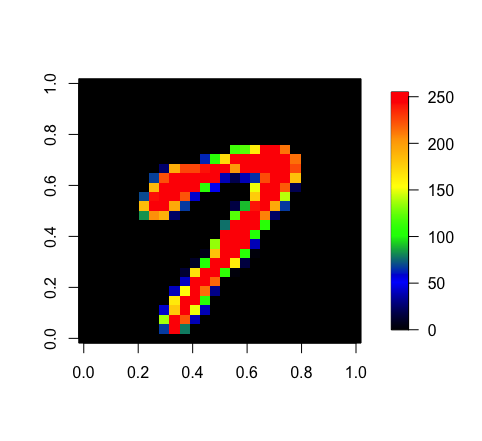}
\includegraphics[width = 0.18\textwidth]{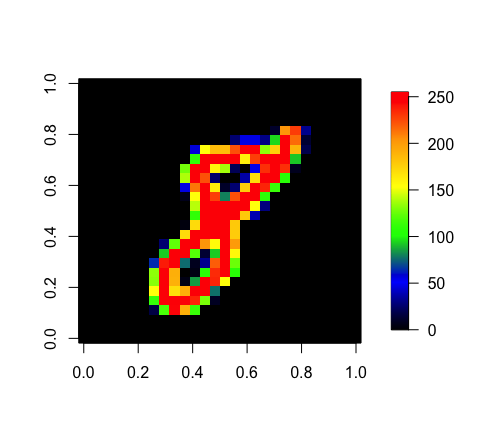} 
\includegraphics[width = 0.18\textwidth]{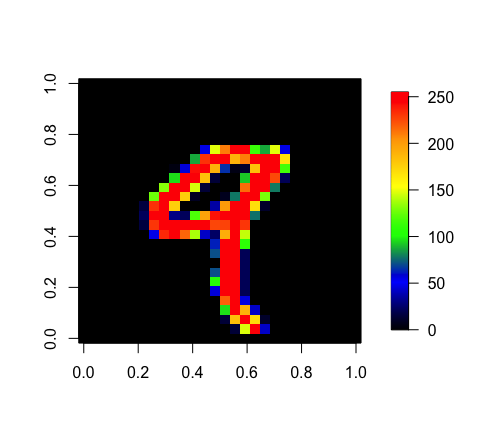}
\\
\end{center}
\end{figure}

\begin{table} 
\caption{\label{TAB:MNIST}Classification accuracy for MNIST data.}
\centering
\begin{tabular}{ccc}
\hline
\hline
\multicolumn{1}{c}{methods} &\multicolumn{1}{c}{training accuracy} & \multicolumn{1}{c}{testing accuracy}\\
\hline
mfDNN &\textbf{0.998} & \textbf{0.983}\\
\hline
MSDA &0.869 & 0.875\\\hline  
PLDA &- & -\\\hline  
\end{tabular}
\end{table}

\subsection{ADNI database}\label{SEC:Real_2D}
The dataset used in the preparation of this article were obtained from the ADNI database (\url{http://adni.loni.usc.edu}). The ADNI is a longitudinal multicenter study designed to develop clinical, imaging, genetic, and biochemical biomarkers for the early detection and tracking of Alzheimer's Disease (AD).
From this database,  we collect PET data from $79$ patients in AD group, $45$ patients in Early Mild Cognitive Impairment (EMCI) group , and $101$ people in Control (CN) group. This  PET  dataset has been spatially normalized and post-processed. These AD patients have three to six times doctor visits and we select the PET scans obtained in the third visits. People in EMCI group only have the second visit, and we select the PET scans obtained in the second visits. For AD group, patients' age ranges from $59$ to $88$ and average age is $76.49$, and there are $33$ females and $46$ males among these $79$ subjects. For EMCI group, patients' age ranges from $57$ to $89$ and average age is $72.33$, and there are $26$ females and $19$ males among these $45$ subjects. For {CN} group, patients' age ranges from $62$ to $87$ and average age is $75.98$, and there are $40$ females and $61$ males among these $101$ subjects.  All scans were reoriented into $79\times 95 \times 68$ voxels, which means each patient has $68$ sliced 2D images with  $79\times 95$ pixels.  For 2D case, it means each subject has $N=79\times 95=7,505$ observed pixels for each selected image slice.   For 3D case, the observed number of voxels for each patient's brain image observation is $N=79\times 95 \times 68=510,340$.   We randomly split the datasets with a $7:3$ ratio in a balanced manner to form the training set and the testing set, with $100$ repetitions.   

We choose  candidates for $(L,J, \bm p, s)$, such that $L=(2,3)^\intercal$, $J=(50,100,200)^\intercal$, $\|\bm p\|_\infty \ = (200,500,800)^\intercal$, and $s=(0.01,0.1,0.5)$ for dropout rate. We still compare our method with MSDA.  For 2D case, it means each subject has $N=79\times 95=7,505$ observed pixels for each selected image slice.   Table \ref{TAB:real_adni_2D} displays the miclassification rates   for  2D   brain imaging data of AD, EMCI and CN. For 3D case, the observed number of voxels for each patient's brain sample is $N=79\times 95 \times 68=510,340$. Unfortunately, given more than a half million covariates, MSDA method crashed down as such gigantic covariance matrix (almost million by million dimension) needs around 2TB RAM to store. It easily  exceeds the memory limit of the  common supercomputer. Hence, MSDA for 3D data results are unavailable. Meanwhile, as PLDA recasts Fisher’s discriminant problem as a biconvex problem that can be optimized using a simple iterative algorithm, PLDA avoids the heavy computation burden of the covariance matrix and it still works for this 3D case. Table \ref{TAB:real_adni_3D} presents the empirical misclassification risk for mfDNN  and PLDA. 

There are several interesting findings in Tables \ref{TAB:real_adni_2D} and \ref{TAB:real_adni_3D}. First,   our proposed classifier   has better performance than other competitors in any  2D slice data or 3D data cases.  
Second, from  Table \ref{TAB:real_adni_3D}, we can conclude that given a single slice of 2D imaging data, the misclassification rates are consistently larger than using the 3D data. It indicates that  3D data contains more helpful information to label the brain images among three stages of the disease. Third, the $10$-th and the $20$-th slices provide the lowest ones among all 2D data. As a matter of fact, it is well known that Alzheimer's disease destroys neurons and their connections in hippocampus, the entorhinal cortex, and the cerebral cortex. The related regions are corresponding to the first $25$ slices. This is a promising finding for neurologists, as this smallest risk indicates this particular slice presents useful information to distinguish the CN, EMCI and AD groups. Further medical checkups are meaningful for this special location in the brain.

\begin{figure}
\begin{center}
\hspace{.4cm}\textbf{AD} \hspace{2.9cm}\textbf{EMCI}\hspace{2.9cm}\textbf{CN}\\
$20$-th
$\begin{array}{l}
\includegraphics[width = 0.28\textwidth]{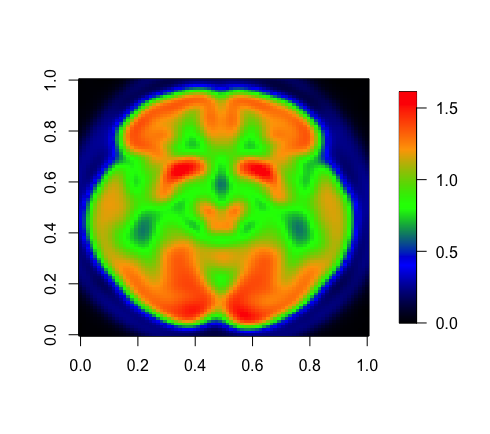} 
\hspace{1.mm}
\includegraphics[width = 0.28\textwidth]{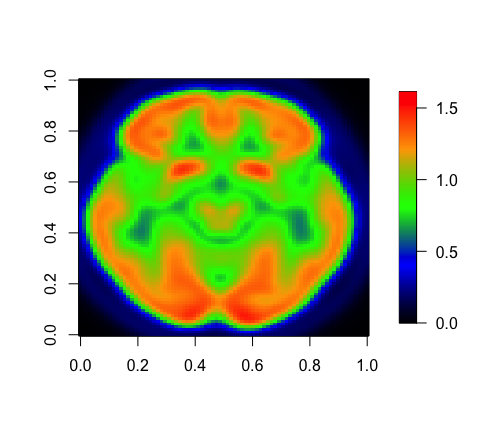} 
\hspace{1.mm}
\includegraphics[width = 0.28\textwidth]{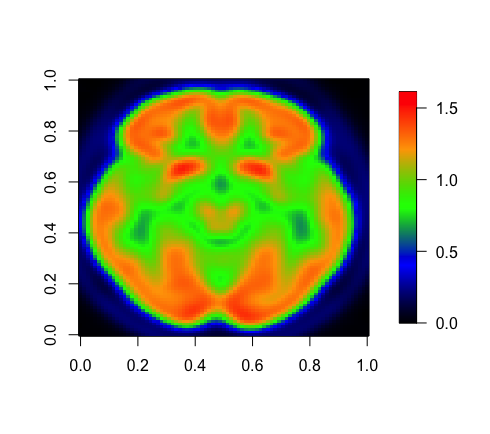} \\
\end{array}$\\
\vspace{-.6cm}
$40$-th
$\begin{array}{l}
\includegraphics[width = 0.28\textwidth]{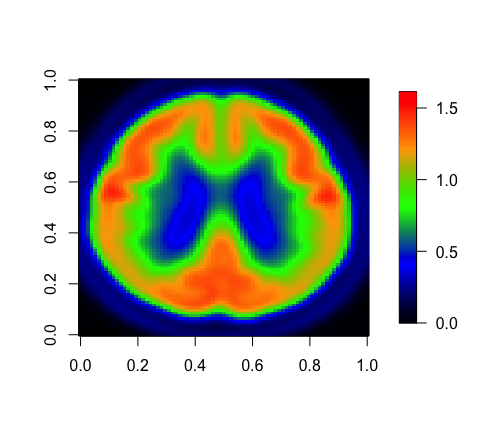} 
\hspace{1.mm}
 \includegraphics[width = 0.28\textwidth]{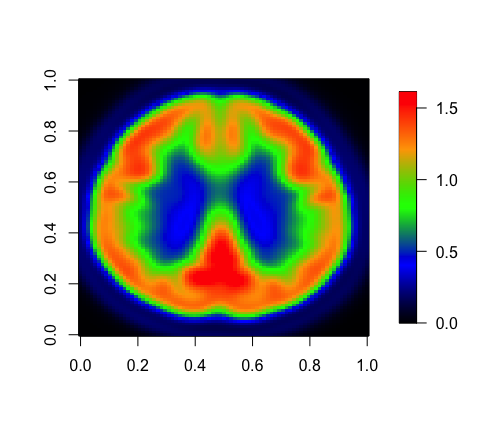} 
 \hspace{1.mm}
 \includegraphics[width = 0.28\textwidth]{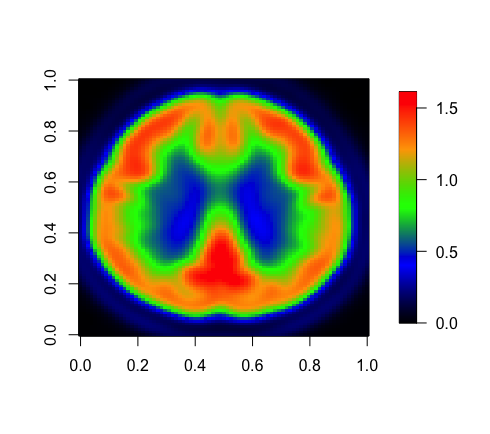} \\
\end{array}$\\
\vspace{-.6cm}
$60$-th
$\begin{array}{l}
\includegraphics[width = 0.28\textwidth]{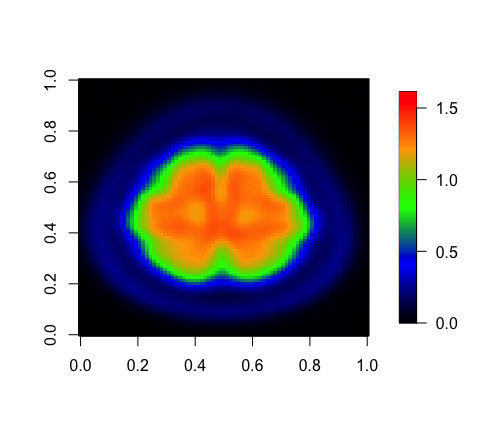} 
\hspace{1.mm}
 \includegraphics[width = 0.28\textwidth]{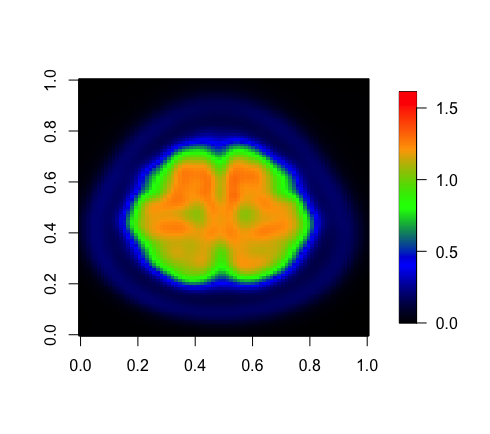} 
 \hspace{1.mm}
 \includegraphics[width = 0.28\textwidth]{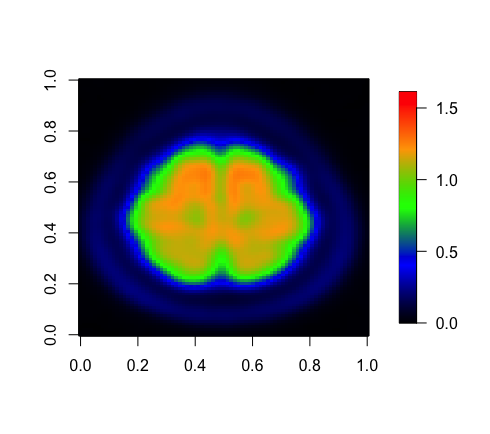} \\
\end{array}$\\
\vspace{-.6cm}
\caption{\label{fig:ADNI_2d} Averaged images of the the $20$-th, the $40$-th and the $60$-th slices of EMCI (left column) group and AD  group (right column).} 
\end{center}
\end{figure}

\begin{table}
\caption{\label{TAB:real_adni_2D}Averaged misclassification rates with standard errors in brackets for ADNI 2D brain images.}
\centering
\begin{tabular}{@{\extracolsep{0.1pt}} ccccccc}
\hline
\hline   
methods&$10$-th & $20$-th & $30$-th & $40$-th & $50$-th & $60$-th \\ \hline
mfDNN &\textbf{0.341} &\textbf{0.325} &\textbf{0.409} &\textbf{0.359} &\textbf{0.365} &\textbf{0.452}  \\
& {(0.057)}& {(0.049)}& {(0.049)}& {(0.050)}& {(0.041)}& (0.055)\\ \hline
MSDA &{0.347} &0.374 &0.444 &0.366 &0.385 &{0.453}   \\
& {(0.038)}& (0.034)& (0.035)& (0.034)& (0.039)& {(0.040)}  \\ \hline
PLDA &0.354 &{0.339} &0.440 &0.376 &0.424 &0.474   \\
& (0.057)& {(0.059)}& (0.067)& (0.060)& (0.068)& (0.064)  \\ \hline
\end{tabular}
\end{table}

\begin{table} 
\caption{\label{TAB:real_adni_3D} Averaged misclassification rates with standard errors in brackets for ADNI 3D brain images.}
\centering
\begin{tabular}{@{\extracolsep{0.1pt}} cc}
\hline
\hline   
\multirow{1}{*}{methods}&  Misclassification rates  \\
  \hline
mfDNN &  \textbf{0.274 (0.044)}\\\hline
MSDA  & -- (--)   \\\hline
PLDA  & 0.295 (0.056)   \\\hline
\end{tabular}
\end{table}


\section{Summary} \label{SEC:Summary}
In this paper, we  propose a new formulation to derive   multiclass  classifiers for multidimensional functional data. We show that our proposal has a solid theoretical foundation and can be solved by
a very efficient computational algorithm. Our proposal actually gives a unified treatment of  both  one-dimensional and multidimensional classification problems.
In light of this evidence, our proposal is regarded as an efficient multiclass and generalization of the  multiclassification methods from i.i.d. data to multidimensional functional data cases.  To the best of our knowledge, the present work is the first work on multiclassification for multidimensional functional data   with   theoretical justification.

\section*{Acknowledgements}

ADNI data used in preparation of this article were obtained from the Alzheimers Disease Neuroimaging Initiative (ADNI) database (\url{adni.loni.usc.edu}). As such, the investigators within the ADNI contributed to the design and implementation of
ADNI and/or provided data but did not participate in analysis or writing of this report.
A complete listing of ADNI investigators can be found at: \url{http://adni.loni.usc.edu/wp-content/uploads/how_to_apply/ADNI_Acknowledgement_List.pdf}.

\bibliographystyle{plain}
\bibliography{Ref}  
\end{document}